%% file: main.tex
\documentclass[journal]{IEEEtran}

\usepackage{amsmath,amsfonts}
\usepackage{algorithmic}
\usepackage{algorithm}
\usepackage{array}
\usepackage[caption=false,font=normalsize,labelfont=sf,textfont=sf]{subfig}
\usepackage{textcomp}
\usepackage{stfloats}
\usepackage{url}
\usepackage{verbatim}
\usepackage{graphicx}
\usepackage{cite}
\usepackage{booktabs}
\usepackage{multirow}
\usepackage{arydshln}
\usepackage{graphicx}

\newcommand{\specialcell}[2][c]{%
  \begin{tabular}[#1]{@{}c@{}}#2\end{tabular}}

\begin{document}

\title{Adaptive Parametric Activation: Unifying and Generalising Activation Functions Across Tasks}

\author{Konstantinos Panagiotis Alexandridis$^{1,*}$, Jiankang Deng$^{2}$, Anh Nguyen$^{3}$  and Shan Luo$^{4}$
\thanks{Manuscript received: 15th October, 2025.}

\thanks{$^{1}$K. P. Alexandridis is with Huawei Noah's Ark Lab, London,  United Kingdom. E-mail: \tt\small\ konstantinos.alexandridis@huawei.com.}%

\thanks{$^{2}$J. Deng is with Department of Computing, Imperial College London, London SW7 2AZ, United Kingdom. E-mail: \tt\small j.deng16@ic.ac.uk.}%

\thanks{$^{3}$A. Nguyen is with the Department of Computer Science, University of Liverpool, Liverpool L69 3BX, United Kingdom. E-mail: \tt\small\ anh.nguyen@liverpool.ac.uk.}

\thanks{$^{1}$S. Luo is with Department of Engineering, King's College London, London WC2R 2LS, United Kingdom. E-mail: \tt\small\ shan.luo@kcl.ac.uk.}%

\thanks{$^{*}$Corresponding author.}%
}

\markboth{Preprint under review.}%
{Shell \MakeLowercase{\textit{et al.}}: A Sample Article Using IEEEtran.cls for IEEE Journals}


\maketitle

\begin{abstract}
  The activation function plays a crucial role in model optimisation, yet the optimal choice remains unclear. For example, the Sigmoid activation is the de-facto activation in balanced classification tasks, however, in imbalanced classification, it proves inappropriate due to bias towards frequent classes.  In this work, we delve deeper in this phenomenon by performing a comprehensive statistical analysis in the classification and intermediate layers of both balanced and imbalanced networks and we empirically show that aligning the activation function with the data distribution, enhances the performance in both balanced and imbalanced tasks. To this end, we propose the
 Adaptive Parametric Activation (APA) function, a novel and versatile activation function that unifies most common activation functions under a single formula. APA can be applied in both intermediate layers and attention layers, significantly outperforming the state-of-the-art on several imbalanced benchmarks such as ImageNet-LT, iNaturalist2018, Places-LT, CIFAR100-LT and LVIS. Also, we extend APA to a plethora of other tasks such as classification, detection, visual instruction following tasks, image generation and next-text-token prediction benchmarks. APA increases the performance in multiple benchmarks across various model architectures. The code is available at {\bf \url{ https://github.com/kostas1515/AGLU}}.
\end{abstract}


\input{sections/introduction}
\input{sections/related_work}
\input{sections/preliminary}
\input{sections/method}
\input{sections/experiments}

\input{sections/conclusion}

\bibliographystyle{IEEEtran}
\bibliography{ref}

\appendix

\section{Representations' Quality}
We evaluate the quality of the representations learned by the SE and APA* models using the recently proposed Neural Collapse framework \cite{papyan2020prevalence}.

Let $f_{k,j} \in \mathrm{R^d}$ be the features of the penultimate layer, $k=\{1,2,...,K\}$ the class, $n_k$ the  number of samples in the class $k$ and $n=\sum_{k=1}^K n_k$ the total number of samples in the dataset. Then the global feature $f_G$ and class prototype $\bar{f_k}$ are:
\begin{equation}
    f_G = \frac{1}{n}\sum_{k=1}^K\sum_{j=1}^{n_k} f_{k,j} \;, \bar{f_k} = \frac{1}{n_k}\sum_{j=1}^{n_k} f_{k,j}
\end{equation}
The within-class covariance matrix $\Sigma_W \in \mathrm{R^{d \times d}}$ and between-class covariance matrix $\Sigma_B \in \mathrm{R^{d \times d}}$ are:
\begin{equation}
\begin{split}
    \Sigma_W :=& \frac{1}{n}\sum_{k=1}^K\sum_{j=1}^{n_k} (f_{k,j} - \bar{f_k})(f_{k,j} - \bar{f_k})^{\top}\\
    \Sigma_b :=& \frac{1}{K}\sum_{k=1}^K (\bar{f_k} - f_G)(\bar{f_k} - f_G)^{\top}\\
\end{split}
\end{equation}
The $\Sigma_W$ matrix shows how distant are individual features $f_{k,j}$ from their class prototype $\bar{f_k}$ and it is an indicator of feature compactness. The $\Sigma_b$ matrix shows how distant are the class prototypes from the global feature, indicating the class separability. Using these matrices we measure the Neural collapse Variability $NC1$ according to \cite{zhu2021geometric} as follows:
\begin{equation}
    NC1 := \frac{1}{K} trace(\Sigma_W \Sigma_b ^\dagger)
    \label{eq:neural_collapse_nc1}
\end{equation}
where the $\dagger$ symbol denotes the pseudo inverse of $\Sigma_b$. $NC1$ measures the magnitude of the
within-class covariance $\Sigma_W$ compared to the magnitude of the between-class covariance $\Sigma_b$ as explained in \cite{zhu2021geometric}.

In practise, a low $NC1$ measure shows that the model has more compact features since $\Sigma_W \downarrow$ decreases and more separable class prototypes because the $\Sigma_b \uparrow$ increases. 
Having more compact features and more separable class prototypes make the representations better and enhance the classification results as shown empirically in previous works \cite{Zhong_2023_CVPR,yang2022inducing,yang2023neural,xie2023neural}.

Using Equation \ref{eq:neural_collapse_nc1}, we measure the $NC1$ of the deep features of the penultimate layer of SE and APA*, in Table \ref{tab:neural_collapse_results} using ImageNet-LT test-set. As the results suggest, our APA* has lower $NC1$ measure for all backbones, showing that APA* produces superior representations that are more compact and seperable than the baseline. This provides another qualitative explanation why our APA* has better performance than SE.
\begin{table}[htb]
    \centering
    \caption{Neural Collapse NC1 measure, on ImageNet-LT test set. APA* has lower $NC1$ measure than the baseline, which indicates that it has learned superior representations.}
    \begin{tabular}{c|c|c}
    \hline
         Backbone&SE-$NC1\downarrow$&APA*-$NC1\downarrow$  \\ \hline
         ResNet-50&3.04&\textbf{2.71}\\
         ResNeXt-50&3.38&\textbf{2.55}\\ 
         ResNet-101&3.15&\textbf{2.69}\\ 
         ResNet-152&3.24&\textbf{2.69}\\ 
    \hline
    \end{tabular}
    
    \label{tab:neural_collapse_results}
\end{table}

\section{Implementation Details}
The implementation details of APA* and AGLU are shown in Table \ref{tab:implementation_details}. For balanced ImageNet-1K, the $\lambda$ parameters  are initialised as random variables drawn from a Uniform distribution (U), with low parameter $0$, and high parameter $1.0$. The APA $\kappa$ parameters are initialised with $U(0,1)$ and the AGLU $\kappa$ parameters are with initialised with $U(1,1.3)$. For all other downstream tasks, that use a pretrained model, such as COCO, LVIS, Places-LT and V3Det, we don't re-initialise the $\kappa$ and  $\lambda$ parameters and we simply load them from the pretained ImageNet1K model. We do not use weight decay for the $\kappa$ and  $\lambda$ parameters as in \cite{he2015delving}.

\subsection{Stable APA implementation}
During the development of APA, we found that it is more stable to use Softplus $s_f(z,\beta) = \frac{1}{\beta} \ln(1 + \exp(\beta z))$, than double exponents, when computing APA. Thus our stable code implementation is:
\begin{equation}
    \eta_{ad}(z,\kappa,\lambda) = \exp(\frac{1}{\lambda} s_f(\kappa z -\ln(\lambda),-1))
\end{equation}
and it is equivalent to the APA used in the main paper.

\begin{table*}[t]
    \centering
    \caption{Implementation details for Long-tailed Datasets, across various architectures.}
    \begin{tabular}{c|c|c|c|c|c}
    \hline
         \multirow{2}{*}{Method}&\multicolumn{1}{c|}{ImageNet-LT}&i-Naturalist18&Places-LT&C100-LT&LVISv1\\
         &R50/X50&R50&R152&R32&MRCNN-R50\\
         \hline
         Batch size&256&1024&256&512&16\\
         Optimiser&SGD&SGD&SGD&SGD&SGD\\
         LR&0.2&0.5&0.1&0.2&0.02\\
         epochs&200&500&40&500&24\\
         Weight Decay&1e-4&1e-4&5e-5&1e-3&1e-4\\
         Norm Weight Decay&1e-4&0.0&5e-5&1e-3&1e-4\\
         Bias Weight Decay&0.0&0.0&0.0&0.0&0.0\\
         Attention Dropout&0.1&0.0&0.1&0.1&0.1\\
         Mixup $\alpha$&0.2&0.2&0.2&0.2&-\\
         CutMix $\alpha$&-&1.0&1.0&-&-\\
         Label smoothing $\epsilon$&-&0.1&0.1&-&-\\
         Repeated Aug&-&\checkmark&-&-&-\\
         AutoAugment&\checkmark&-&\checkmark&\checkmark&-\\
         RandAugment&-&\checkmark&-&-&-\\
         Erasing prob&-&0.1&-&-&-\\
         Cutout&-&-&-&\checkmark&-\\
         Cos. Cls. scale&16&16&learnable&learnable&N/A\\
         Norm. Mask scale&N/A&N/A&N/A&N/A&learnable\\
         Sampler&random&random&random&random&RFS\\
         APA $\kappa$ Init&U(-1,0)&U(0,1)&N/A&U(-1,0)&N/A\\
         APA $\lambda$ Init&U(0,1)&U(0,1)&N/A&U(0,1)&N/A\\
         AGLU $\kappa$ Init&U(1,1.3)&U(1,1.3)&N/A&U(1,1.3)&N/A\\
         AGLU $\lambda$ Init&U(0,1)&U(0,1)&N/A&U(0,1)&N/A\\
         \hline
    \end{tabular}

    \label{tab:implementation_details}
\end{table*}

\subsection{AGLU derivatives}
\noindent \textit{Proof of Eq. 9}. Then the gradient of AGLU with respect to $\kappa$ is:
\begin{equation}
    \begin{aligned}
        \frac{\partial AGLU(x,\kappa,\lambda)}{\partial \kappa}
        = \partial\frac{ x \cdot (\lambda \exp(-\kappa x)+1)^{\frac{1}{-\lambda}}}{\partial \kappa} \\
        = x \frac{(\lambda \exp(-\kappa x) +1)^{(\frac{-1}{\lambda}-1)}}{-\lambda} \cdot (-\lambda x \exp(-\kappa x)) \\
        = x^2 \exp(-\kappa x) \frac{(\lambda \exp(-\kappa x) +1)^{(\frac{-1}{\lambda})}}{\lambda \exp(-\kappa x) +1}\\
        = x^2 \frac{\eta_{\text{ad}}(x,\lambda,\kappa)}{\lambda +\exp(\kappa x)}\\
    \end{aligned}
\end{equation}

\noindent \textit{Proof of Eq. 10}. Then the gradient of AGLU with respect to $\lambda$ is:
\begin{equation}
    \begin{aligned}
        \frac{\partial AGLU(x,\kappa,\lambda)}{\partial \lambda}
        &= \partial\frac{ x \cdot (\lambda \exp(-\kappa x)+1)^{\frac{1}{-\lambda}}}{\partial \lambda} \\
        &= x \frac{(\lambda \exp(-\kappa x) +1)^{( \frac{-1}{\lambda} -1)}}{-\lambda} \cdot (\exp(-\kappa x)) \\
        &= \frac{-x}{\lambda} \exp(-\kappa x) \frac{(\lambda \exp(-\kappa x) +1)^{(-\frac{1}{\lambda})}}{\lambda \exp(-\kappa x) +1}\\
        &= \frac{-x}{\lambda} \frac{\eta_{\text{ad}}(x,\lambda,\kappa)}{\lambda +\exp(\kappa x)}\\
    \end{aligned}
\end{equation}

\noindent \textit{Proof of Eq. 11}. Then the gradient of AGLU with respect to $\lambda$ is:
    \begin{multline}
        \frac{\partial AGLU(x,\kappa,\lambda)}{\partial x}
        = \partial\frac{ x \cdot (\lambda \exp(-\kappa x)+1)^{\frac{1}{-\lambda}}}{\partial x} \\
        =\eta_{\text{ad}}(x,\lambda,\kappa) + x \cdot \partial\frac{  (\lambda \exp(-\kappa x)+1)^{\frac{1}{-\lambda}}}{\partial x} \\
        = \eta_{\text{ad}}(x,\lambda,\kappa) + x \frac{(\lambda \exp(-\kappa x) +1)^{( \frac{-1}{\lambda} -1)}}{-\lambda} \cdot (-\kappa \lambda \exp(-\kappa x)) \\
        =\eta_{\text{ad}}(x,\lambda,\kappa) + \kappa x \exp(-\kappa x) \frac{(\lambda \exp(-\kappa x) +1)^{(-\frac{1}{\lambda})}}{\lambda \exp(-\kappa x) +1}\\
        = \eta_{\text{ad}}(x,\lambda,\kappa) +  \kappa x  \frac{\eta_{\text{ad}}(x,\lambda,\kappa)}{\lambda +\exp(\kappa x)}\\
    \end{multline}

\section{Results}
\subsection{Impact of Initialisation}
In all long-tail experiments, we have initialised $\lambda$ using the Uniform distribution with low parameter 0 and high parameter 1 as a default. Regarding the $\kappa$ parameter inside AGLU, we initialise it to be close to $1.0$, as this works best, as shown in Table \ref{tab:aglu_init_res}. Regarding the $\kappa$ parameter inside the attention layer, we found that initialising it with $Uniform(-1,0)$ is slightly better than $Uniform(0,1.0)$ as shown in Table \ref{tab:apastar_init_res}. In all other experiments including image generation, ImageNet21K, ImageNet1K training, VLM training and GPT training we used $\kappa$ and $\lambda$ initialization from the $Uniform(0.8,1.2)$ distribution, to simplify the codebase. The latter initialization has good performance across many balanced tasks. We note that it might be possible to achieve better performance by tuning the initialisation for every task, however we chose not to do it due to computing constrains. 

\begin{table}[htb]
    \centering
    \caption{AGLU-$\kappa$ parameter initialisation, using APA* ResNet50 backbone on ImageNet-LT. The $\lambda$ is initialised with $Uniform (0,1)$ by default.}
    \begin{tabular}{c|c}
    \hline
    AGLU - $\kappa$&top-1 \\ 
    \hline
    $Uniform(0,1)$&57.7\\
    $Uniform(-2,0)$&57.3\\
    $Uniform(-3,0)$&57.4\\
    $Uniform(-2,2)$&Failed\\
    $Uniform(1,1.3)$&\textbf{57.9} \\
    \hline
    \end{tabular}
    \label{tab:aglu_init_res}
\end{table}

\begin{table}[htb]
    \centering
    \caption{$\kappa$ parameter initialisation inside the attention layer, using APA* ResNet50 backbone on ImageNet-LT. The $\lambda$ is initialised with $Uniform (0,1)$ by default.}
    \begin{tabular}{c|c}
    \hline
    APA - $\kappa$&top-1 \\ 
    \hline
    $Uniform(0,1)$&57.6\\
    $Uniform(-1,0)$&\textbf{57.9}\\
    \hline
    \end{tabular}
    \label{tab:apastar_init_res}
\end{table}

\subsection{Channel specific $\lambda$ and $\kappa$}
We have also tried a variant that uses seperate $\lambda$ and $\kappa$ parameters for every channel. As shown in Table \ref{tab:channel_specific_res}, this variant performs worse than using shared  $\lambda$ and $\kappa$ parameters for the channels.  
\begin{table}[htb]
    \centering
    \caption{Results with Channel Specific $\lambda$ and $\kappa$, using APA* ResNet50 backbone on ImageNet-LT.}
    \begin{tabular}{c|c}
    \hline
    APA &top-1 \\ 
    \hline
    Channel Specific&57.5\\
    Channel Shared&\textbf{57.9}\\
    \hline
    \end{tabular}
    \label{tab:channel_specific_res}
\end{table}

\subsection{Baseline enhancements}
We show the detailed ablation study for ImageNet-LT in Table \ref{tab:base_enhancements}. First, the vanilla ResNet50 model trained for 100 epochs on ImageNet-LT achieves $44.4\%$. When we train for 200 epochs then it adds $1.5$pp and switching from linear classifier to cosine classifier adds another $0.4$pp. Stronger training techniques such Mixup~\cite{zhang2017mixup}, Auto-Augment~\cite{cubuk2019autoaugment} and weight decay tuning further boost the performance by $3.3$pp. Post-calibrated Softmax \cite{hong2021disentangling} adds an additional $5.4$pp and finally the Squeeze and Excite module \cite{hu2018squeeze} adds another $1.0$pp reaching the final $56.0\%$. Most baseline performance comes from the PC-Softmax and the weight decay finetuning. On top of this strong baseline, our APA increases the performance by $1.0$pp, showing its strong generalisability. Dropout and LayerNorm further increase the performance by $0.4$pp and finally AGLU adds a respectable $0.5$pp reaching $57.9\%$ accuracy on ImageNet-LT. The absolute improvement of all modules is $13.5$pp and our proposed methods, APA and AGLU, contribute by $1.5$pp which is a relative $11\%$ of the total absolute improvement.  

\begin{table*}[t]
    \centering
    \caption{Detailed Ablation Study, using ResNet50 on ImageNet-LT.}
    \begin{tabular}{p{0.5cm}p{0.5cm}p{0.5cm}p{0.5cm}p{0.5cm}p{0.5cm}p{0.5cm}p{0.5cm}p{0.5cm}p{0.5cm}p{0.5cm}|c}
         \rotatebox[origin=c]{270}{200 epochs}&
         \rotatebox[origin=c]{270}{Cosine Classifier}&
         \rotatebox[origin=c]{270}{SE-nets\cite{hu2018squeeze}}&\rotatebox[origin=c]{270}{AutoAugment\cite{cubuk2019autoaugment}}
         &\rotatebox[origin=c]{270}{Mixup \cite{zhang2017mixup}}&\rotatebox[origin=c]{270}{Weight Decay Tuning \cite{alshammari2022long}}
         &\rotatebox[origin=c]{270}{PCS \cite{hong2021disentangling}}&\rotatebox[origin=c]{270}{APA (ours)}
         &\rotatebox[origin=c]{270}{Dropout \cite{hinton2012improving}}&\rotatebox[origin=c]{270}{LayerNorm \cite{ba2016layer}}
         &\rotatebox[origin=c]{270}{AGLU (ours)}
         &ImagetNet-LT\\
         \hline
         &&&&&&&&&&&44.4\\
         \hline
         \checkmark&&&&&& &&&&&45.9 \\
         \checkmark&\checkmark&&&&& &&&&&46.3\\
         \checkmark&\checkmark&\checkmark&&&& &&&&&46.8\\
         \checkmark&&&\checkmark&&& &&&&&45.2\\
         \checkmark&&\checkmark&\checkmark&&& &&&&&45.9\\
         \checkmark&\checkmark&\checkmark&\checkmark&&&&&&&&46.6\\
          \checkmark&\checkmark&\checkmark&\checkmark&\checkmark&&&&&&&46.6\\
         \checkmark&\checkmark& &\checkmark&\checkmark&\checkmark &&&&&&49.6\\
         \checkmark&\checkmark& &\checkmark&\checkmark&\checkmark&\checkmark&&&&&55.0\\
         \checkmark&\checkmark&\checkmark&\checkmark&\checkmark&\checkmark &&&&&&51.7\\
         \checkmark&\checkmark&\checkmark&\checkmark&\checkmark&\checkmark&\checkmark&&&&&56.0\\
         \hline
         \checkmark&\checkmark&\checkmark&\checkmark&\checkmark&\checkmark&\checkmark&\checkmark&&&&57.0\\
        \checkmark&\checkmark&\checkmark&\checkmark&\checkmark&\checkmark&\checkmark&\checkmark&\checkmark&&&57.3\\
       \checkmark&\checkmark&\checkmark&\checkmark&\checkmark&\checkmark&\checkmark& \checkmark&\checkmark&\checkmark& &57.4\\
        \checkmark&\checkmark&\checkmark&\checkmark&\checkmark&\checkmark&\checkmark&\checkmark&\checkmark&\checkmark&\checkmark&\textbf{57.9}\\
    \end{tabular}
    
    \label{tab:base_enhancements}
\end{table*}

\subsection{Qualitative Results}
\begin{figure}[h]
    \centering
    \includegraphics[width=1\linewidth]{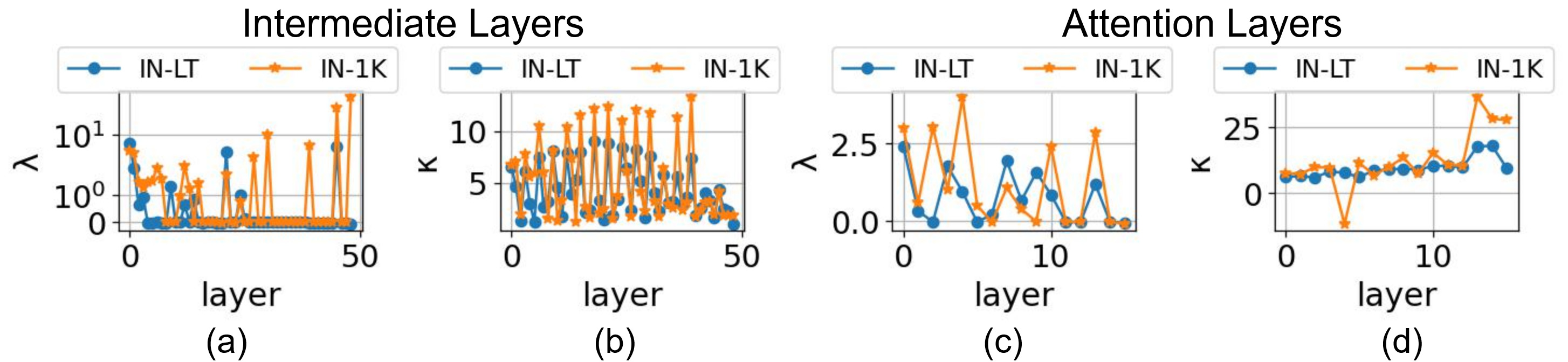}
    \caption{Visualisations of the learned $\lambda$ and $\kappa$ parameters for balanced ImageNet1K (IN-IK) training in blue, and imbalanced ImageNet-LT (IN-LT) training in blue.}
    \label{fig:lambda_kappa_in1k_ilt}
\end{figure}

In Figure~\ref{fig:lambda_kappa_in1k_ilt}, we show the learned parameters, when training with the balanced and imbalanced ImageNet. Regarding the $\lambda$ inside the AGLU layers in (a),  we see that both balanced-trained and imbalanced-trained networks prefer an all-pass filter for the early 2-3 layers, possibly, because the networks are uncertain which features to remove. Then in the intermediate layers, we observe smaller $\lambda$ that corresponds to harder filters and in the final semantic layers we observe larger $\lambda$, possibly, because the network prefers smoother semantic features in order to have smoother classification boundaries. 
In (b), we see a `down-down-up' $\kappa$-pattern in most bottlenecks, for both balanced and imbalanced ImageNet, showing that the networks prefer softer activations, at first, and harder activations before the residual connection. This indicates that the networks, first, keep most information inside the bottleneck's projections, and second, they disregard any redundant information, using harder activation, only before performing addition using the residual connection. 

Finally, in the last bottlenecks, i.e. layers 45-50, the $\kappa$ parameter diminishes, showing that the network prefers overly smooth activations, possibly, to enhance the classification using smoother classification boundaries.

Regarding the attention layers, the $\kappa$ parameter in (d) dominates over the influence of $\lambda$ in (c), showing that hard channel attention is more preferable than soft channel attention for all layers. 

\subsubsection{Visualisations on Imagenet-LT}
We further show more qualitative results on ImageNet-LT with ResNet50 backbone. On the left subfigure, we show the model's highest prediction marked with F,C,R that stands for frequent, common and rare class respectively and the Grad-cam activation~\cite{selvaraju2017grad}. On the right subfigure, we show the last layer's channel attention signal and its corresponding entropy denoted with (E). As the Figure shows,  APA* produces higher entropy attention signals than the baseline and predicts both frequent and rare classes correctly. 
\begin{figure*}
    \centering
    \includegraphics[width=1\linewidth]{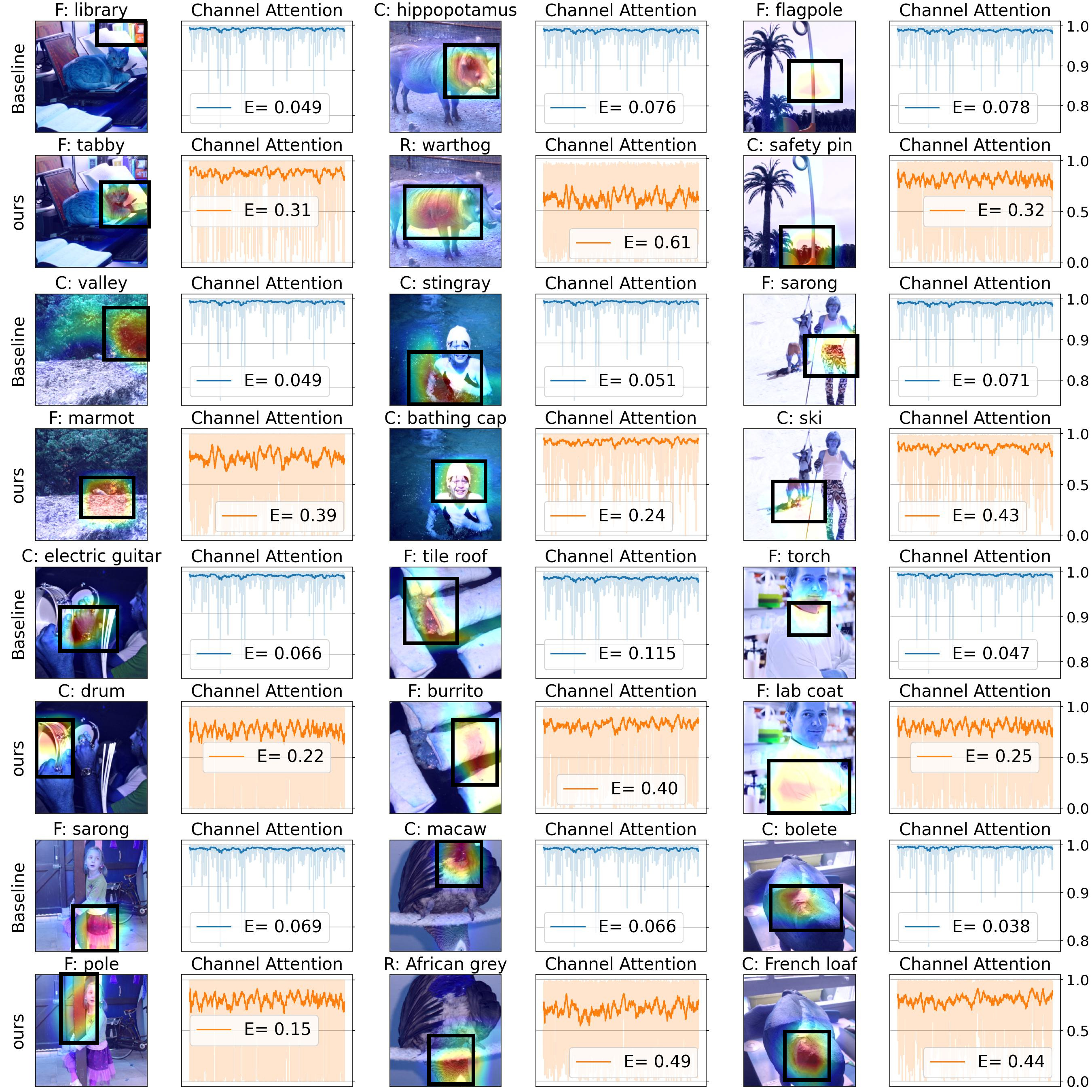}
    \caption{Comparative Results between the SE-ResNet50 (baseline) and APA*-ResNet50 (ours) with respect to the activations (left) and the attention entropy (right). F,C,R denote frequent, common and rare samples from ImageNet-LT. Our method produces attention signals that have significantly larger entropy than the baseline for both frequent and rare classes.  }
    \label{fig:qualitative_results_with_box_supp}
\end{figure*}

\begin{figure*}[htb]
    \centering
    \includegraphics[width=1\linewidth]{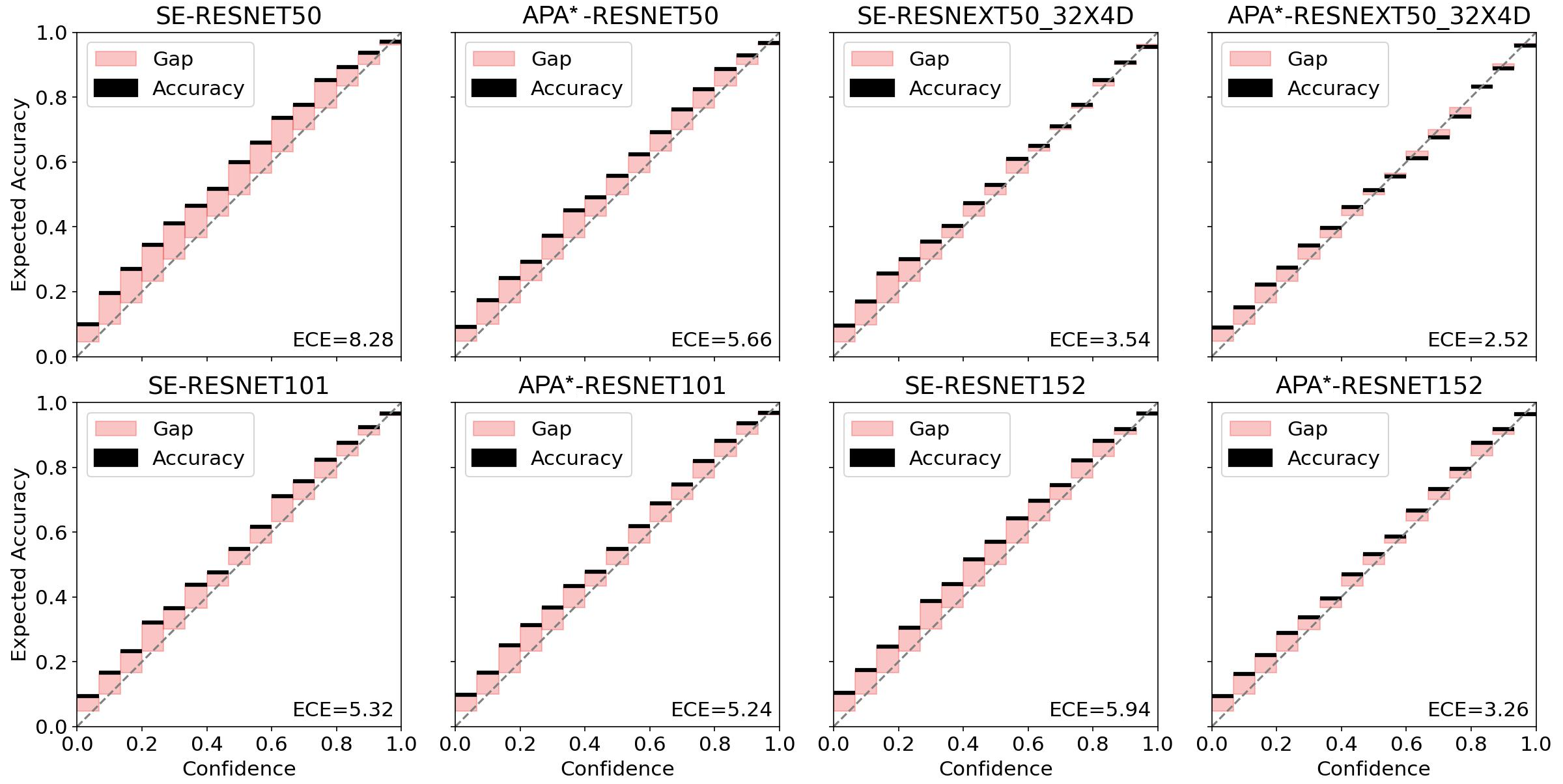}
    \caption{Calibration results using ResNets on ImageNet-LT. SE (left) is underconfident, i.e., its confidence scores are lower than its actually accuracy due to over-regularisation. Our APA* (right) reduces the ECE and makes more accurate predictions with higher confidence than SE. }
    \label{fig:calibration_results}
\end{figure*}
\subsection{Calibration results}
Calibration is an important property of models, since it reassures that the confidence of the prediction matches the actual accuracy. When models are not calibrated, then they give wrong predictions with high confidence score (overconfident models) or make correct predictions with low confidence score (underconfident models). In both situations, the miscalibrated models cannot help in the decision making process because their predictions do not reflect their actual accuracy.

In practice in long-tailed learning, the use of complex augmentations and regularisations like mixup, cutmix, label-smoothing, auto-augment and cosine classifier may improve the accuracy but it also reduces the confidence of the model due to over regularisation.
As shown in Figure \ref{fig:calibration_results} (left-subfigure), SE-Resnet50 is under-confident due to the usage of complex training that includes heavy augmentations and regularisations. 
When APA* is applied, it reduces the Expected Calibration Error (ECE) as shown in Figure \ref{fig:calibration_results} (right-subfigure) for all backbones.

\end{document}

%% file: sections/introduction.tex
\section{Introduction}
\label{sec:intro}
\IEEEPARstart{I}{mage} recognition has witnessed tremendous progress over the last years due to the use of deep learning, large image datasets such as ImageNet1K \cite{deng2009imagenet} and advances in model architectures \cite{dosovitskiy2021an,he2016deep}, learning algorithms \cite{he2022masked,radford2021learning,he2020momentum}, activation layers \cite{hendrycks2016gaussian,glorot2011deep,he2015delving} and normalisation techniques \cite{ba2016layer,ioffe2015batch}. In this work, we focus on the activation layer of the network.

 In balanced image classification works, it was empirically shown that if the activation function is close to the real data distribution then the model converges faster because the learning objective becomes easier \cite{he2015delving,hendrycks2016gaussian}. Based on this, the GELU~\cite{hendrycks2016gaussian} and the PRELU~\cite{he2015delving} were proposed as alternatives to the commonly used RELU~\cite{glorot2011deep} and they were utilised inside the model's layers to activate the intermediate features of the network. Similarly, in imbalanced image classification, many works have empirically shown that the Sigmoid or the Softmax activation functions, are inappropriate and using another activation function increases the performance~\cite{Ren2020balms,hong2021disentangling,menon2021longtail,alexandridis2022long}. Based on that, the Gumbel activation \cite{alexandridis2022long} and the Balanced Softmax~\cite{Ren2020balms} activation were proposed and they were used inside the classification layer to predict the classes.  In contrast to balanced classification, these works focused only on the classification layer and they disregarded the importance of the intermediate activations.  To this end, there is no principled way of choosing the right activation function, and usually, practitioners use the best activation function, according to the task, through cross-validation or parameter-tuning.

In this work, we focus on this problem. First, we theoretically show that the activation function enforces a prior belief of how the data is distributed and therefore it acts as an initialisation point. In practise, a good initialisation point enhances the convergence, therefore, having an appropriate activation function can increase the performance.
Second, we study the impact of the activation function on balanced and imbalanced classification from two perspectives, i.e. the classification layer, and the intermediate layers.
\begin{figure}[t]
    \centering
    \includegraphics[width=1\linewidth]{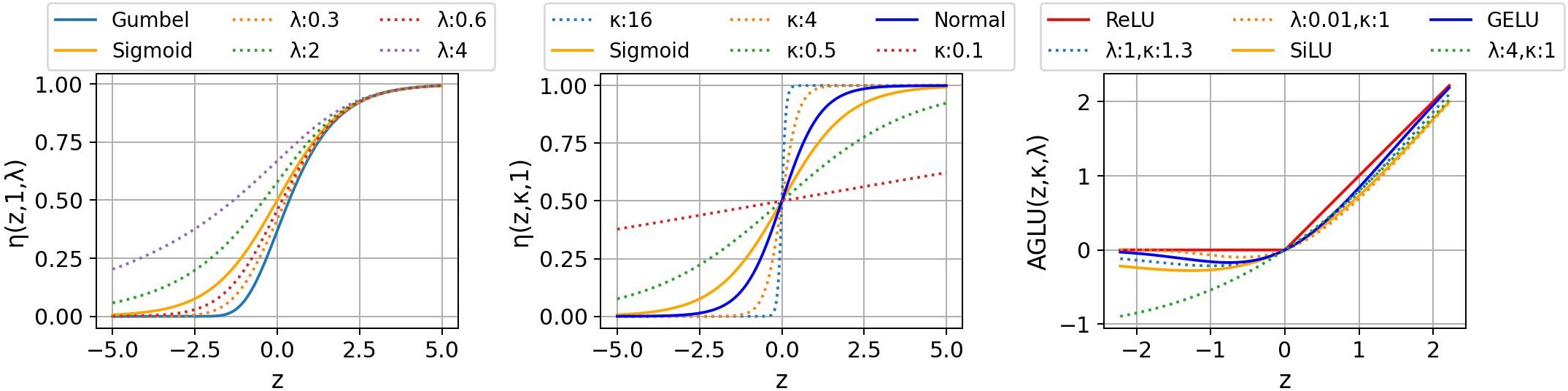}
    \caption{Our APA unifies most activation functions under the same formula.}
    \label{fig:unified_activations}
\end{figure}

 Our findings show that the classification logit distribution of a pretrained model heavily depends on the degree of data imbalance. For example, in balanced training, the classification logits align better with the Logistic distribution, while in imbalanced learning they align better with the Gumbel distribution. Regarding the intermediate layers, we study the channel attention layer as an example and we find that it is also affected by the degree of data imbalance. We find that, in balanced learning, the channel attention is robust for all classes, however in imbalanced training the channel attention enhances more the frequent classes than the rare classes.

To this end, we empirically show that the commonly used Sigmoid activation function cannot generalise for both balanced and imbalanced learning, because it is non-parametric and does not align with the imbalanced data distribution.
Motivated by this, we develop a novel Adaptive Parametric Activation (APA) function shown in Figure \ref{fig:unified_activations}. APA allows the model to align its activations to both the balanced and imbalanced data distributions and reach great performance in both tasks. 
APA has several benefits, it unifies most previous activations functions such as the Sigmoid, the Gumbel \cite{alexandridis2022long}, the RELU \cite{glorot2011deep}, the SiLU \cite{hendrycks2016gaussian} and the GELU\cite{hendrycks2016gaussian} under a common formula. Also, it uses two learnable parameters that allow the network to select the best activation function during optimisation enlarging the model's capacity.
Moreover, our APA is versatile, it can be used as a direct replacement to RELU, or it can replace the Sigmoid activation function inside the attention mechanism boosting the performance significantly and consistently. Finally, APA can be generalised to both imbalanced and balanced tasks and surpass the state-of-the-art (SOTA).
Our \textbf{contributions} are:
\begin{itemize}
    \item We demonstrate the importance of the activation function in balanced and imbalanced data distributions, through statistical analysis;
    \item We propose the novel APA function that unifies most common activation functions under a single formula;
    \item We have validated the efficacy of APA on a range of long-tailed benchmarks, including ImageNet-LT \cite{liu2019large}, iNaturalist18 \cite{van2018inaturalist}, Places-LT \cite{liu2019large}, CIFAR100-LT \cite{cao2019learning}, LVIS \cite{gupta2019lvis} largely surpassing the state of the art.
    \item We comprehensively evaluate APA across various architectures and benchmarks, including image classification, object detection, VQA tasks, image generation and text classification and show that APA generalises to both large scale datasets and foundation models.
\end{itemize}

%% file: sections/related_work.tex
\section{Related Work}
\label{sec:related_work}

\subsection{Long-tailed image recognition.}
Long-tailed image recognition can be grouped according to representation learning and classifier learning techniques. Representation learning techniques improve the feature quality through rare class features generation~\cite{vigneswaran2021feature,wang2021rsg,zang2021fasa}, contrastive objectives~\cite{wang2021contrastive,cui2021parametric,zhu2022balanced,samuel2021distributional,li2022targeted,kang2021exploring}, ensemble or fusion models~\cite{wang2021long,li2022nested,9774921,zhou2020bbn,li2022trustworthy,cui2022reslt,cai2021ace}, knowledge distillation~\cite{li2022nested,iscen2021cbd,he2021distilling,li2021self}, knowledge transfer~\cite{liu2019large,parisot2022long,zhu2020inflated} and data augmentation~\cite{zhong2021improving,park2022majority,xu2021towards}. 
Classifier learning techniques enhance rare class classification through decoupled training~\cite{kang2019decoupling,zhang2021distribution,wang2020devil,kim2020adjusting,hsu2023abc}, margin adjustment~\cite{menon2021longtail,Ren2020balms,hong2021disentangling,cao2019learning,hyun2022long,pan2021model,wang2021seesaw,wang2022c2am,zhao2022adaptive,alexandridis2023inverse,alexandridis2025fractal}, cost-sensitive learning~\cite{cui2019class,khan2017cost,wang2017learning,huang2016learning}, resampling~\cite{chawla2002smote,mahajan2018exploring,shen2016relay,zou2018unsupervised,gupta2019lvis,park2022majority,hong2022safa,zang2021fasa}, dynamic loss adaptation based on batch statistics~\cite{hsieh2021droploss,tan2020equalization,wang2021adaptive}, gradient statistics~\cite{tan2021equalization,li2022equalized}, weight norms~\cite{wang2022c2am} and classification scores~\cite{feng2021exploring,he2022relieving}.
These techniques are efficient, however they are applied either during the input phase or during the loss disregarding the intermediate activations.

\subsection{Attention Networks.}
Attention networks, such as spatial attention~\cite{jaderberg2015spatial}, channel attention~\cite{hu2018squeeze} and spatial-channel attention~\cite{woo2018cbam}, are widely used in image recognition. Self-attention is the core mechanism of the Visual Transformer (ViT) architecture~\cite{dosovitskiy2021an}, and recently many improvements have been made to the Transformer architecture~\cite{touvron2021training,touvron2021going,liu2021swin,xiao2021early} and its training procedure~\cite{steiner2021train,beyer2022better,touvron2022deit}. 

APA can be combined with many attention models like channel attention, Spatial, Spatial-Channel attention and self-attention models and further boost the performance.

\subsection{Activation functions}
The RELU function~\cite{glorot2011deep} dominates the landscape of deep image classification, and it is especially used inside the convolutional networks, while the GELU and SiLU \cite{hendrycks2016gaussian} are more commonly used inside the transformer network \cite{dosovitskiy2021an} or the ConNext models \cite{liu2022convnet,woo2023convnext}. The Gumbel activation is effective for imbalanced image recognition tasks \cite{alexandridis2022long}, and recently, it showed good performance for balanced tasks as well \cite{das2025gompertz}. The PRELU function~\cite{he2015delving} is a generalisation of the RELU, because it linearly activates both positive and negative inputs and the ELU~\cite{clevert2015fast} is a follow-up work that enforces saturation on the negative inputs after some threshold. Compared to those, our APA, generalises the RELU, the GELU, SiLU and Gumbel and it can also be seen as a smoother version of PRELU. Recently, there has been increased interest in diffeomorphic activation functions used in image classification \cite{chelly2024trainable}, segmentation \cite{freifeld2017transformations}, parameter-efficient tuning \cite{mantri2024digraf} and graph neural networks \cite{mantri2024digraf}. However, these activation functions require extensive forward and backward computations and a user defined input-domain interval. In contrast, APA forward and backward is straightforward and it operates in the whole input domain. 

%% file: sections/preliminary.tex
\section{Preliminaries}
\subsection{Activation function.}
First, we show the importance of the activation function in model optimisation, following \cite{alexandridis2022long}.
Let's consider the example of binary classification, where $z$ is the input, $y \in \{0,1 \}$ is the target class and $f(z)=W^T z+b$ is the classification network. The target class and the input $z$ are related as:
\begin{equation}
y=
    \begin{cases}
        1, \;\; \text{if} \;\;  f(z) + \epsilon>0\\
        0, \;\; \text{otherwise}
    \end{cases}
\end{equation}
where $\epsilon$ is the error, that is a random variable and it is distributed according to the real data distribution. In this example, the classification boundary is set to 0, however, it can be adjusted by the network's bias $b$ during optimisation.
The probability of the class $P(y=1)$ is:
\begin{equation}
\begin{aligned}
P(y=1)=P(f(z) + \epsilon>0)=P(\epsilon>-f(z))\\
=1-F(-f(z))
\label{eq:cumulative}
\end{aligned}
\end{equation}
where $F$ is the cumulative distribution function.
During network optimisation, if one chooses the Sigmoid activation function $\sigma(z) = (1+\exp(-z))^{-1}$, then the prediction $\bar{y}$ is obtained using $\bar{y}= \sigma(f(z))$ and the probability $P(\bar{y}=1)$ is:

\begin{equation}
\begin{aligned}
P(\bar{y}=1)= \frac{1}{1+\exp(-f(z))} = F_{\text{logistic}}(f(z);0,1) \\
= 1-F_{\text{logistic}}(-(f(z);0,1)
\label{eq:log_prior}
\end{aligned}
\end{equation}
Comparing Eq.~\ref{eq:cumulative} and Eq. \ref{eq:log_prior}, it is shown that when we use the Sigmoid, we assume that the error term $\epsilon$ follows the standard Logistic distribution. If we use Gumbel activation then we assume that the error follows the Gumbel distribution \cite{alexandridis2022long} and, in general, any activation function assumes a different distribution of $\epsilon$.

For this reason, the activation function can be seen as an initialisation point, or a prior belief of how the real data are distributed. If the prior is good, then the learning objective becomes easier and the performance is increased as it was shown empirically by many past studies \cite{he2015delving,alexandridis2022long,glorot2011deep,skorski2021revisiting}.

\subsection{Balanced versus Imbalanced learning}

In this subsection, we perform a statistical analysis on the empirical distributions of the logits and the intermediate activations, to understand the importance of the activation function in balanced and imbalanced learning.

\begin{figure}[t]
    \centering
    \includegraphics[width=1\linewidth]{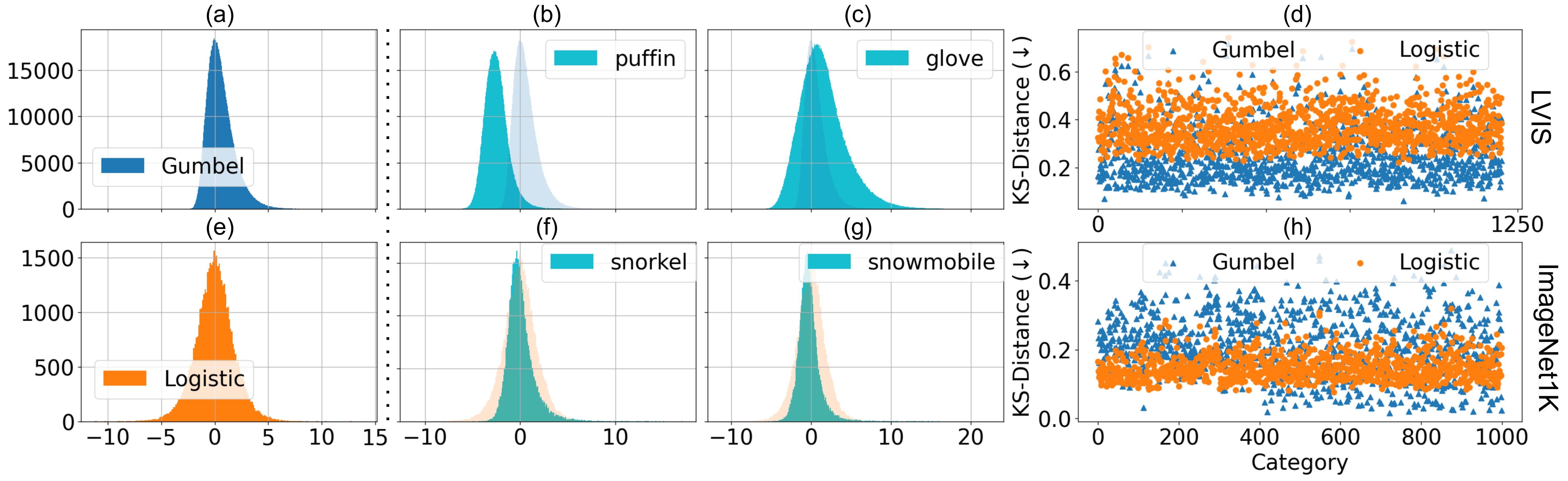}
    \caption{Top: In imbalanced learning, the logit distributions are more skewed and they have a smaller KS distance to the Gumbel than the Logistic distribution as shown in (d). Bottom: In balanced learning, the logit distributions are less skewed and they align better with the Logistic, than the Gumbel distribution, as shown in (h).}
    \label{fig:logits_lvis_imnet}
\end{figure}

\noindent \textbf{Classification logits.}
We train a MaskRCNN \cite{he2017mask} ResNet50\cite{he2016deep} on LVIS dataset \cite{gupta2019lvis}, which is a highly imbalanced object detection dataset containing 1203 classes. After the model has converged, we perform inference and store the predicted classification logits, i.e., $f(z_y)$ for every class $y$. Next, we perform histogram binning on the logits and we visualise the empirical distributions for the rare class \textit{puffin} and the frequent class \textit{glove} in Fig.~\ref{fig:logits_lvis_imnet} (b) and (c) respectively. 

The rare class has a negative average logit value, because it is dominated by the frequent classes, which have higher average logit values as shown in (c). Regarding its distribution, it resembles the Gumbel distribution, because it has a heavy right tail and it is skewed. To quantify that, we calculate the statistical distance, using the Kolmogorov-Smirnov (KS) test \cite{massey1951kolmogorov}, between all empirical class distributions and the Gumbel and Logistic theoretical distributions, shown in (a) and (e). As shown in (d), most classes have smaller distance to the Gumbel distribution.
We repeat this test, using ResNet50 trained on the balanced ImageNet1K. As shown in (f) and (g), the classification logits are different this time, for example most average logit values are centered around zero and they are less skewed. In general, the logit distributions are closer to the logistic distribution as shown in (h).
This explains why the Sigmoid activation achieves better performance in the balanced classification task and worse performance in imbalanced classification \cite{alexandridis2022long,Ren2020balms,hong2021disentangling,menon2021longtail}.
Next, we show that data imbalance also affects the intermediate layers, by studying the channel attention as an example.

\noindent \textbf{Attention layer.}
Attention mechanism for input $X \in \mathbb{R}^{H\times W \times C}$ re-weights the input features $X$ by applying an attention function $A(X)$, i.e., $X'=A(X) \otimes X$. For example, in Channel Attention (CA) \cite{hu2018squeeze}, $X'= \sigma(\text{MLP}(\text{GAP}_c(X))) \otimes X$, where $\text{GAP}_c\in \mathbb{R}^{1\times1\times C}$ is Global Average Pooling and $ \otimes$ is the element-wise product. In this case, the attention function is $A_{CA}(X)= \sigma(\text{MLP}(\text{GAP}_c(X)))$. 
\label{sec:prelimineries}

\begin{figure*}[t]
    \centering
    \includegraphics[width=0.9\linewidth]{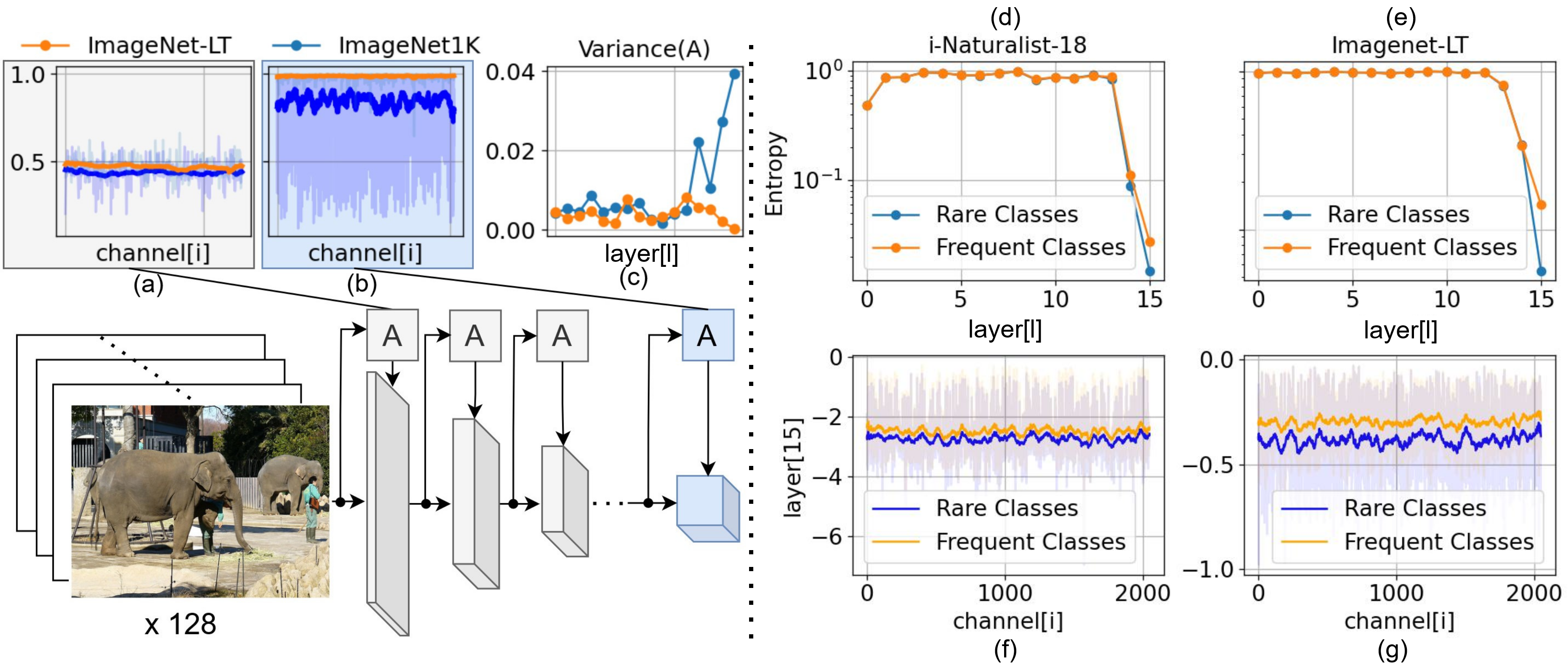}
    \caption{Visualisations of channel attention (A). In (a) the attention signals when training with imbalanced ImageNet and balanced ImageNet have similar variance in the first layer but completely different in the last most semantic layer in (b). The variance with ImageNet-LT training drops to zero for deeper layers as shown in (c), because the attention promotes only the frequent classes. In (d) and (e) the entropy of channel attention is smaller for the rare classes than the frequent classes in both i-Naturalist18 and ImageNet-LT training. In (f) and (g) the channel responses are smaller for the rare classes forboth i-Naturalist18 and ImageNet-LT. These observations shows that channel attention is not robust during imbalanced training, motivating the development of APA. }
    \label{fig:channel_attention}
\end{figure*}

\noindent \textbf{Balanced vs Imbalanced channel attention.}
\label{subsec:channel_imbalance}
We train SE-ResNet50~\cite{hu2018squeeze} models (SE-R50) on balanced ImageNet-1K and imbalanced ImageNet-LT. After training the models, we analyse the average channel attention signals for a random batch of 128 test images. Fig. \ref{fig:channel_attention}-a shows the output $A_{CA}$, in the first layer of SE-R50, Fig.~\ref{fig:channel_attention}-b shows the output, in the last layer, and Fig.~\ref{fig:channel_attention}-c shows the variance of $A_{CA}$ across all layers. 
As shown in Fig. \ref{fig:channel_attention}-a, the attention is similar in the first layer and it is different in the last layer, as shown in Fig. \ref{fig:channel_attention}-b. 
As displayed in Fig. \ref{fig:channel_attention}-c, the attention variance with ImageNet1K (blue-curve) is larger and the attention reweights all channels and affects all classes. In contrast, for the imbalanced case (orange-curve), the attention signal has small variance, indicating that it is biased to some classes.

\noindent \textbf{Layer-wise analysis.}
This phenomenon is most prevalent in the last attention layer, which is the most semantic.  
To quantify which attention layers are affected the most, we use entropy as a measure of signal complexity. Since the channel attention produces a probabilistic weighting vector via the Sigmoid activation, we calculate the total entropy of channel attention, for a layer $l$, as the sum of the binary channel distribution entropies as follows:
\begin{equation}
\begin{split}
    E_l = -\frac{1}{C}\sum_{i=1}^C[(A_{CA}(X_{l})\log(A_{CA}(X_{l}))\\ 
    +(1-A_{CA}(X_{l})\log(1-A_{CA}(X_{l}))]
\end{split}
\end{equation}
When the layer's entropy is closer to zero, the channel attention signals are closer to 1 and they do not affect the original features for that layer, i.e., $X' = 1 \otimes X$.
If the layer's attention entropy is closer to one, then the channel attention signals are informative, as they affect the signal i.e., $X' = A(X) \otimes X$.

To investigate the complexity of the attention signal, we propagate two batches of 64 test images that contain only frequent and only rare classes respectively, through the pretrained SE-R50 and measure the attention entropy of $A_{CA}$.
In Fig.~\ref{fig:channel_attention} (d) and (e), the average entropy is similar for both frequent and rare classes for all layers except for the last layer which is the most semantic. The blue curve, that corresponds to rare class channel attention, has lower entropy than the orange curve that corresponds to frequent class channel attention in the last layer for both i-Naturalist-18 and ImageNet-LT. This indicates that channel attention produces simpler signals for the rare classes and more complex signals for the frequent classes. Finally, in (f) and (g), we show that the average channel responses are smaller when the inputs are rare classes, than frequent classes, which explains why the network cannot model the rare classes effectively.
In conclusion, this analysis shows that data imbalance affects the quality of the activations inside the intermediate layers and it highlights the limitation of the Sigmoid activation to model the rare classes. 

%% file: sections/method.tex
\section{Method}
\subsection{Adaptive Parametric Activation}
As shown in the previous section, the degree of data imbalance affects both the classification logits and the intermediate layers. While it is possible to perform a statistical analysis and select the appropriate activation for the classification layer, this is difficult to do for all layers of the network, because first, the data distributions inside the layers dynamically change during training ~\cite{ba2016layer,ioffe2015batch}, and secondly, there is no one-to-one correspondence between classes and intermediate channels, which hinders attribution. For this reason, we propose the Adaptive Parametric Activation (APA):
\begin{equation}
    \begin{split}
    \eta_{ad}(z,\kappa,\lambda) = (\lambda \exp(-\kappa z)+1)^{\frac{1}{-\lambda}}\\
    \end{split}.
    \label{eq:adaptive_activation}
\end{equation} 
APA can adjust its activation rate, dynamically, according to the input's distribution using two parameters $\kappa$ and $\lambda$, that can be learned during optimisation. $\kappa \in \mathbb{R}$ is the gain parameter that controls the function's sensitivity.
$\lambda \in (0,\infty)$ is the asymmetrical parameter that controls the function's response rate to positive and negative inputs, allowing the model to have different learning degrees when the input is positive or negative.    
This function is also known as Richard's curve \cite{richards1959flexible} and it unifies the most common activation functions.

For example, if $\kappa=\lambda=1$ then APA becomes the Sigmoid activation that has a symmetric response rate for both positive and negative inputs
and it is successful for balanced classification tasks. 
If $\kappa=1,\lambda \xrightarrow[]{} 0$ then APA becomes the Gumbel activation that has an asymmetric response rate and it is successful for long-tailed instance segmentation tasks \cite{alexandridis2022long}. This behaviour is shown in Fig. \ref{fig:unified_activations} left and middle.
Based on Eq. \ref{eq:adaptive_activation}, we also define the Adaptive Generalised Linear Unit (AGLU):
\begin{equation}
    \begin{split}
     AGLU(z,\kappa,\lambda) =z \cdot \eta_{ad}(z,\kappa,\lambda) \\
    \end{split}
    \label{eq:agelu}
\end{equation}

\begin{figure}[t]
    \centering
    \includegraphics[width=1\linewidth]{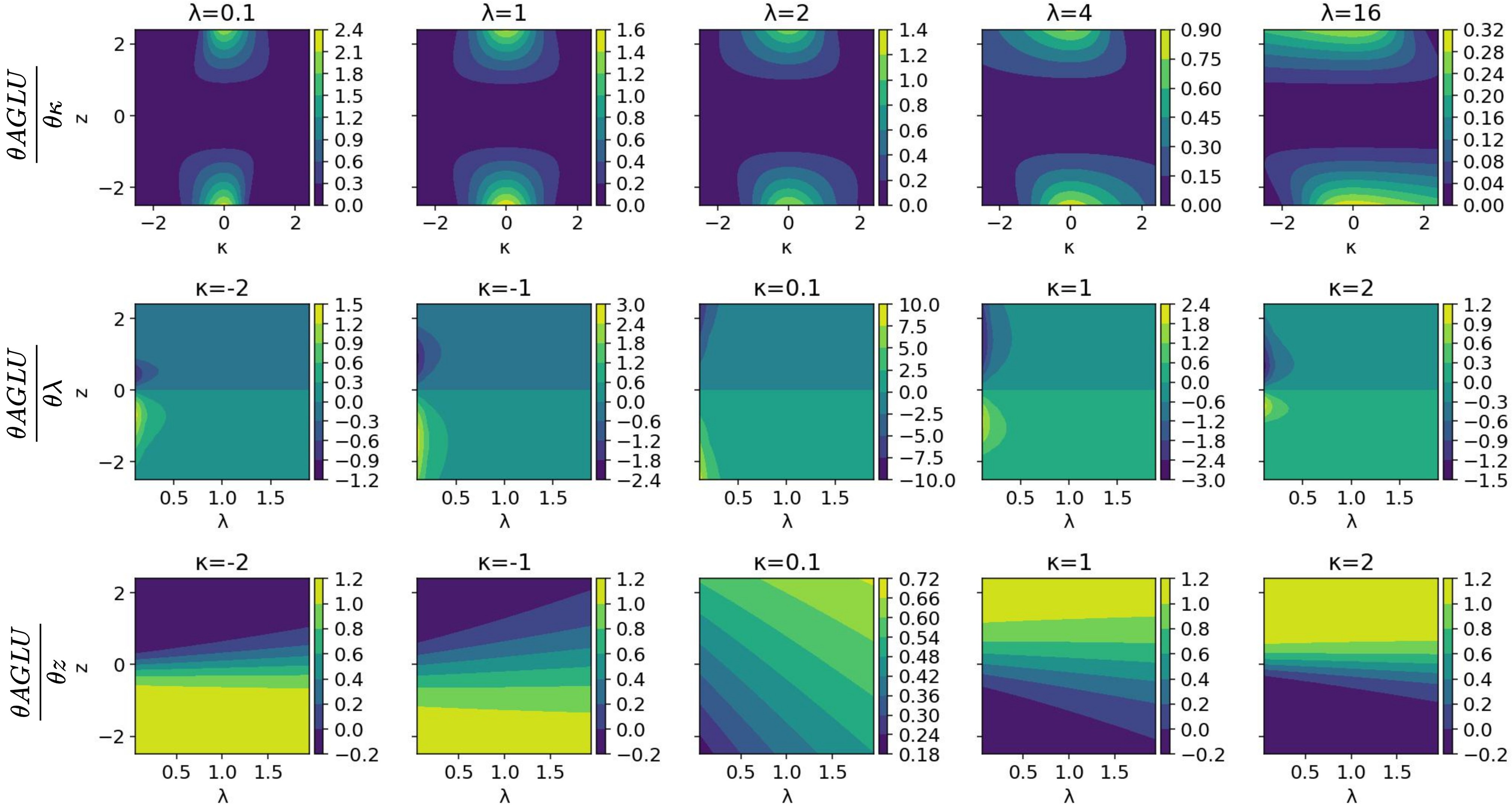}
    \caption{AGLU derivatives with respect to $\kappa$ (top), $\lambda$ (middle) and $z$ (bottom).}
    \label{fig:aglu_derivatives}
\end{figure}

AGLU has many interesting properties, for example, if $\kappa=\lambda=1$ then AGLU becomes the Sigmoid Linear Unit (SiLU)~\cite{hendrycks2016gaussian}. If $\kappa=1.702,\lambda=1$, then it becomes the Gaussian Error Linear Unit (GELU)~\cite{hendrycks2016gaussian}.  If $\kappa \xrightarrow[]{} \infty$, then AGLU becomes ReLU\cite{glorot2011deep} and if $\lambda \xrightarrow[]{} \infty$, then AGLU becomes the identity function, as shown in Fig. \ref{fig:unified_activations}-right. 

In other words, the $\kappa$ parameter controls the RELU-ness and the $\lambda$ parameter controls the leakage. Also, AGLU could be seen as a smoother version of PRELU, because $AGLU(z,1,\lambda \xrightarrow[]{} \infty)= PRELU(z,1)$. We compare AGLU to most common activation functions, in more detail, in Table \ref{tab:aglu_comp}. The derivative of AGLU with respect to $\kappa$ is:
\begin{equation}
    \frac{\partial AGLU(z,\kappa,\lambda)}{\partial \kappa} = z^2 \cdot \frac{\eta_{ad}(z,\kappa,\lambda)}{\lambda + \exp(\kappa z)}
\end{equation}
the derivative of AGLU with respect to $\lambda$ is:
\begin{equation}
    \frac{\partial AGLU(z,\kappa,\lambda)}{\partial \lambda} = -\frac{z}{\lambda} \cdot \frac{\eta_{ad}(z,\kappa,\lambda)}{\lambda + \exp(\kappa z)}
\end{equation}
and the derivative of AGLU with respect to $z$ is:
\begin{equation}
    \frac{\partial AGLU(z,\kappa,\lambda)}{\partial z} =  \kappa z \cdot\frac{\eta_{ad}(z,\kappa,\lambda)}{\lambda + \exp(\kappa z)} + \eta_{ad}(z,\kappa,\lambda)
\end{equation}
The proofs of the derivatives are shown in the Appendix.
The derivatives of AGLU are shown in Fig. \ref{fig:aglu_derivatives}. Using various $\kappa$ and $\lambda$ combinations AGLU has drastically different behaviour and this enhances the capacity of the network and achieves good performance as shown in the experiments. 

\begin{table}[t]
    \centering
    \caption{Comparison of different activation functions.}
    \resizebox{1\linewidth}{!}{%
    \begin{tabular}{c|c|c}
    \hline
         Name&Formula&Output Range  \\
         \hline
         RELU \cite{glorot2011deep}& $\eta(z) = \max(0,z) $&$(0,\infty)$ \\
         Gaussian Unit \cite{hendrycks2016gaussian}& $\eta(z) = z \sigma(1.702z) $& $(-0.17,\infty)$\\
         Sigmoid Unit \cite{hendrycks2016gaussian}& $\eta(z) = z \sigma(z) $ & $(-0.28,\infty)$ \\
         Mish \cite{misra2019mish} & $\eta(z) = z \tanh (\ln(1+\exp(z)) $ & $(-0.31,\infty)$ \\
         PRELU \cite{he2015delving}& $\eta(z,\kappa) = \max(0,z) + \kappa \min(0,z) $& $(-\infty,\infty)$ \\  
         ELU \cite{clevert2015fast} & $\eta(z,\kappa) = \max(0,z) + \kappa(\exp(\min(0,z))-1) $&$(-\kappa,\infty)$ \\
         \hline
         AGLU (ours) &$\eta(z,\kappa,\lambda) =z \cdot (\lambda \exp(-\kappa z)+1)^{\frac{1}{-\lambda}} $& $(-\infty,\infty)$ \\
    \hline
    \end{tabular}
    }
    \label{tab:aglu_comp}
\end{table}
APA is versatile and it has many use cases. For example it can be uses in the intermediate layers, i.e. replacing RELU, or inside the attention mechanism, replacing the Sigmoid activation.

\subsection{Non-Linearity and Adaptability} 
\begin{figure}[htb]
    \centering
    \includegraphics[width=0.86\linewidth]{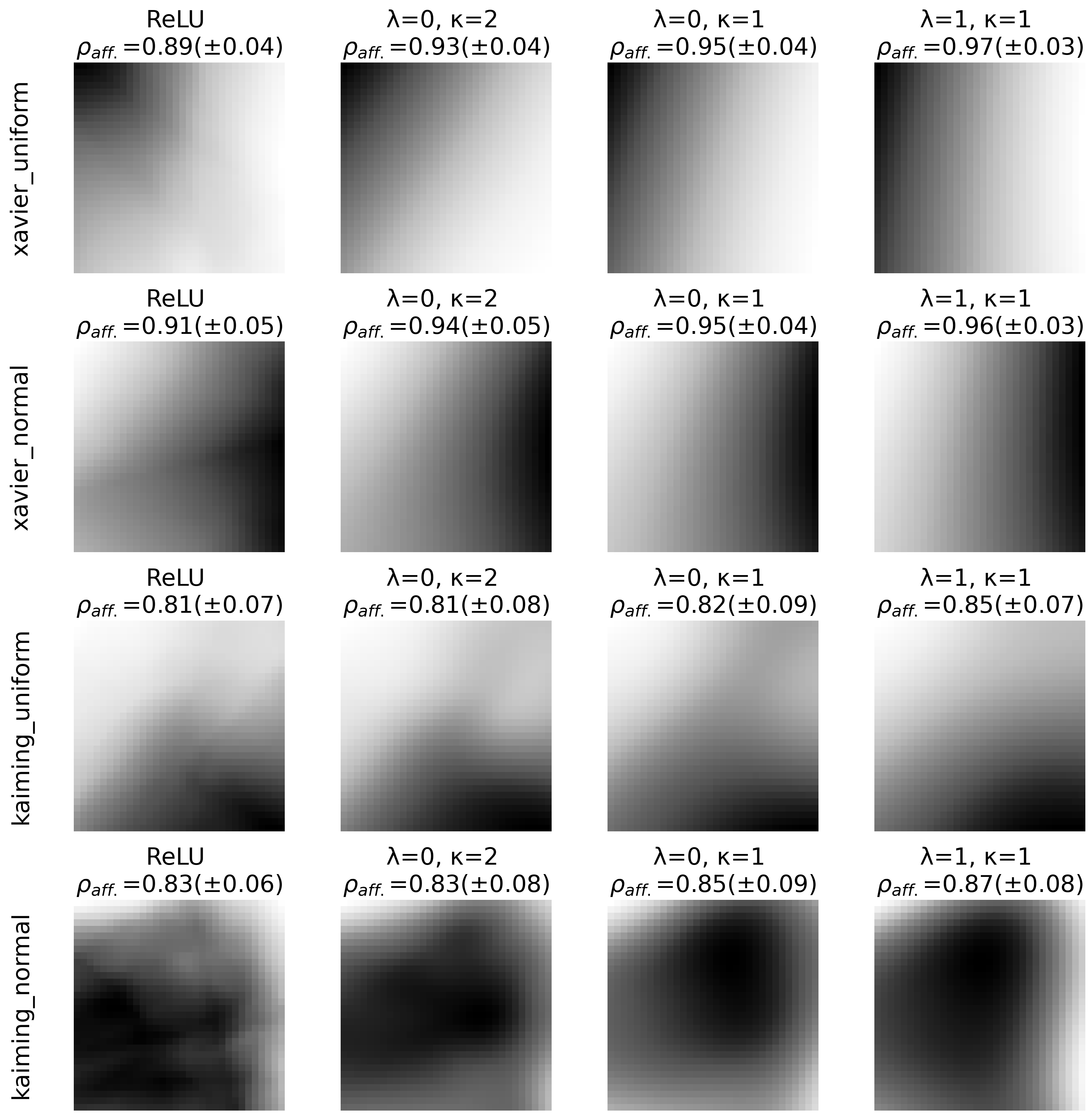}
    \caption{Visualisation of the output of a random MLP, initialised with Xavier-Uniform, Xavier-Normal, Kaiming-Uniform and Kaiming-Normal, across the rows, and $RELU(z)$, $AGLU(z,2,0)$, $AGLU(z,1,0)$ and $AGLU(z,1,1)$ across the columns. The RELU MLP is the most non-linear, for all weight initialisations, because it has the smallest affinity scores $\rho_{aff}$ and most complex outputs as shown in the first column.  Increasing the AGLU's $\kappa$ value lowers $\rho_{aff}$ and makes the MLP more non-linear, as shown in the second column, while decreasing the $\kappa$ or increasing the $\lambda$ values increases $\rho_{aff}$ and makes it more linear as shown in the third and forth columns respectively. }
    \label{fig:implicit_bias}
\end{figure}

APA 's parameters control the amount of non-linearity added to the model. To measure this, we follow the newly proposed Neural Redshift framework \cite{teney2024neural} and use the affinity score $\rho_{aff}$ \cite{bouniot2025alexnet} to study the behavior of a randomly initialised three hidden-layer MLP with a scalar output. We initialise the MLP weights using various schemes such as Xavier \cite{glorot2010understanding} and Kaiming \cite{he2015delving} and the MLP biases with the uniform distribution $U(-1,1)$, we feed an input of $64^2$ evenly spaced 2D coordinates on the $[-1,1]$ grid and visualise the output in grayscale in Figure \ref{fig:implicit_bias}.

If RELU is used as nonlinearity, then the MLP has the biggest amount of non-linearity, because it produces the smallest affinity scores $\rho_{aff}$ and the most complex outputs for all the initialisation schemes as shown in the first column.  
If AGLU with small $\lambda$ and larger $\kappa$ values is used then the AGLU-MLP becomes similar to the RELU-MLP, i.e. it is more non-linear and it has smaller affinity scores $\rho_{aff}$ and more complex outputs as shown in the the second column. Decreasing the $\kappa$ or increasing the $\lambda$ values of AGLU makes the MLP less non-linear, i.e. the affinity scores increase and the outputs become smoother as shown in the third and fourth column respectively. 

This has two effects, it allows the users to smoothly control the inductive bias of the model, i.e. by fixing the amount of non-linearity or it enables to learn the best nonlinearities by optimization.

%% file: sections/experiments.tex
\section{Experiments in Long-tail learning}
\label{sec:exp_setup}
\subsection{Datasets.}
We use CIFAR100-LT~\cite{cao2019learning} with exponential imbalance factor of 100 and 10, ImageNet-LT~\cite{liu2019large}, Places-LT \cite{liu2019large} and iNaturalist2018~\cite{van2018inaturalist} following the common long-tailed classification protocol. We report our results using top-1 accuracy on the balanced test sets, to fairly evaluate all classes. For ImageNet-LT, we split the categories according to their class frequency in the training set, into \textit{Many} ($>$100 images), \textit{Medium} (20$\sim$100 images) and \textit{Low} ($<$20 images) and do per-group evaluation following \cite{liu2019large}. Also, we use the LVISv1~\cite{gupta2019lvis} instance segmentation dataset, which has 100K training images and 1203 classes, that are grouped according to \textit{Frequent} ($>$100 images), \textit{Common} (10$\sim$100 images) and \textit{Rare} ($<$10 images) classes. For this dataset, we report mask average precision $AP$, bounding box average precision $AP^b$ and $AP^r$, $AP^c$ and $AP^f$ which is mask average precision for rare, common and frequent classes respectively.
\input{tables/imagenet_lt_sota}

\subsection{Implementation Details}
We primarily use Squeeze and Excite \cite{hu2018squeeze} as our baseline with ResNet-32 \cite{he2016deep} for CIFAR-LT, ResNet50 for iNaturalist, ResNet50 and ResNext50~\cite{xie2017aggregated} for ImageNet-LT and ResNet152 for Places-LT, which has been pretrained on ImageNet1K according to \cite{liu2019large}.
For LVIS, we use SE-Resnets with MaskRCNN \cite{he2017mask}, FPN~\cite{lin2017feature}, Normalised Mask \cite{wang2021seesaw}, RFS sampler \cite{gupta2019lvis} and GOL loss~\cite{alexandridis2022long} as a baseline. 
 All baselines use bag of tricks \cite{zhang2021bag}, and strong training techniques that we discuss and ablate in the Appendix. Our motivation for using bag of tricks is two-fold, first, it pushes the performance even further and secondly, it showcases the efficacy of our work.

For our attention models, we replace the Sigmoid with APA, and we further use LayerNorm \cite{ba2016layer} and attention dropout~\cite{hinton2012improving} with $p=0.1$, for all datasets except iNaturalist. We denote this configuration as APA* in our Tables and we ablate its components. For our AGLU models, we simply use it in-place of RELU, or GELU. Following \cite{he2015delving} we disable the weight decay of AGLU parameters, allowing the activation curve to reach the optimal shapes during training.

\input{tables/inat_places_cifar_sota}

\subsection{Long-tailed Classification Benchmark.}
\label{sec:results} 
We compare APA* against ensemble and fusion models \cite{wang2021long,cui2022reslt,li2022trustworthy,cai2021ace}, margin adjustment \cite{hong2021disentangling,zhang2021distribution,zhao2022adaptive}, contrastive learning \cite{wang2021contrastive,li2022targeted,kang2021exploring}, knowledge transfer \cite{parisot2022long}, knowledge distillation \cite{he2021distilling,li2021self}, decoupled methods \cite{zhong2021improving,alexandridis2023inverse,zhang2021distribution,hsu2023abc}, sharpness aware minimisation \cite{zhou2023imbsam,zhou2023class,ma2023curvature} and data augmentation \cite{zhong2021improving,park2022majority}.

On ImageNet-LT, as shown in Table \ref{tab:imagenet_sota}, our baseline models with bag of tricks reach the state-of-the-art (SOTA) for both SE-ResNet50 (R50) and SE-ResNeXt50 (X50). We want to point out that most performance comes from the bag of tricks and not the SE module, as shown, in detail in the Appendix. 

Our APA* outperforms the SE-R50 baseline by $1.4$ percentage points (pp) on average, by $1.3$pp on frequent categories, $1.2$pp on medium and $2.2$pp on few classes. 
Most importantly, it increases the performance of both frequent and rare classes, which is a unique advantage compared to the previous works. Additionally, our APA* exceeds RIDE with 3 Experts (3E) and CMO \cite{park2022majority} by $1.2$pp on average and by $3.7$pp on the rare classes using a single model. When AGLU is combined, it pushes the performance of APA* R50, by $0.5$pp on average, by $0.8$pp on the frequent, $0.5$pp on the medium and $0.1$pp on the few classes. 

APA* with X50 increases the SE-X50 baseline by $1.7$pp on average, $1.0$pp the many classes, $1.4$pp the medium classes and $1.7$pp the few classes. 
Furthermore, it outperforms RIDE(3E)+CMO+CR \cite{ma2023curvature} by $1.0$pp on average, $1.6$pp on frequent classes, $0.8$ on common and $1.0$pp on rare classes using a single model. When AGLU is combined to APA* X50, it adds $0.7$pp on average, $0.9$pp on the frequent categories, $0.3$pp on the medium, and $1.7$pp on the few classes. 

\input{tables/ablations}
On iNaturalist18, Places-LT and CIFAR100-LT, our SE baselines with bag of tricks again reach the SOTA, as shown in Table \ref{tab:inat_places}-a and b.

Regarding, iNaturalist18, APA* improves the SE-baseline by $1.0$pp. It also outperforms BCL~\cite{zhu2022balanced} by $0.5$pp,  ResLT by $1.8$pp and LWS+ImbSAM~\cite{zhou2023imbsam} by $1.2$pp. AGLU further enhances the performance by a staggering $3.5$pp compared to SE-baseline, which is a significant increase.

On Places-LT, our APA* also increases the performance of the SE baseline by $0.8$pp. Moreover, it surpasses MisLAS~\cite{zhong2021improving} by $0.9$pp and CC-SAM~\cite{zhou2023class} by $0.7$pp. When AGLU is combined with APA*, it further increases the performance by $0.7$pp reaching $42.0\%$. 
 
As shown in Table \ref{tab:inat_places}-b, on CIFAR100-LT, APA* improves the performance by $0.5$pp and $1.0$pp compared to SE baseline using an imbalance factor of 10 and 100 respectively. Furthermore, compared to RIDE+CMO~\cite{park2022majority}, CC-SAM~\cite{zhou2023class} and RIDE+CMO+CR~\cite{ma2023curvature}, APA* achieves $1.9$pp, $1.1$pp and $1.2$pp higher accuracy respectively, under an imbalance factor of 100 using a single model. Finally AGLU, further boosts the performance of APA* by $1.1$pp and $0.4$pp for imbalance factor of 10 and 100 respectively.

\noindent \textbf{Activation Ablation Study.}
We compare APA, without the Dropout and Layernorm, against different activation functions such as the Sigmoid and the Gumbel~\cite{alexandridis2022long} using the SE-ResNet50 baseline. Also, we implement Sigmoid and Gumbel variants with learnable temperature, to further understand their difference to our adaptive activation. As shown in Table \ref{tab:agca_ablations}-a, Sigmoid with temperature achieves slightly better performance for the rare classes, however it over-fits the frequent categories. In contrast, our APA achieves the best performance, increasing the overall performance by $0.3$pp, the frequent classes by $0.2$pp and the common categories by $0.4$pp compared to the second best Gumbel with temperature.

\noindent \textbf{Generalisation to other Attention mechanisms.}
In Table~\ref{tab:agca_ablations}-b, we show that APA can be combined with other attention mechanisms. Specifically, APA improves the performance of Spatial Attention by $0.4$pp and the Spatial-Channel Attention by $1.3$pp. AGLU further increases the performance of Spatial Attention by $1.2$pp and Spatial-Channel Attention by $0.2$pp.

\input{tables/lvis}

\noindent \textbf{Combining APA* with Classifier Learning.} APA* is an efficient module that can be easily combined with common classifier learning techniques such as margin adjustment \cite{Ren2020balms}, reweighting \cite{cao2019learning} and resampling \cite{kang2019decoupling}. As shown in Table \ref{tab:agca_ablations}-c, APA* consistently boosts the performance of all these methods. 

\noindent \textbf{Larger ResNets.}
We further train deeper models like ResNet-101 and ResNet-152 on ImageNet-LT and compare the SE and APA* attention methods.
As Table~\ref{tab:agca_ablations}-d shows, the APA* module enlarges the performance of all models consistently for both frequent, common and rare classes. Especially, APA* with ResNet101 promotes overall accuracy by $2.1$pp, boosting the rare categories by a significant $4.5$pp compared to SE-Resnet101.

\noindent \textbf{Ablation of APA* components.}
APA* uses Channel attention, APA, Dropout and Layer-Norm and we show their effects in Table~\ref{tab:agca_ablations}-e. The plain R50, without channel attention, achieves $55.0\%$ and SE increases its performance by $1.0$pp. APA increases the performance of the SE baseline by another $1.0$pp. Dropout increases the performance by $0.3$pp and LayerNorm enhances the performance by an additional $0.1$pp. In the Appendix, we show the full component ablation. 

\noindent \textbf{Comparison of AGLU.} In Table~\ref{tab:agca_ablations}-f, we compare AGLU to other commonly used activations functions, using APA* ResNet50 as backbone and switching the activation function of the intermediate layers. Our AGLU outperforms all the other methods and it is the best choice.

\subsection{Long-Tailed Instance Segmentation.}
As the results suggest in Table \ref{tab:lvis1_sota}, the SE-R50 backbone increases the overall mask performance compared to plain GOL-R50 by $0.5$pp but most improvement comes from the common and frequent categories while the rare categories are significantly reduced  by $0.8$pp.
In contrast, our APA* with AGLU-R50 improves the performance by $0.9$pp on average mask and bounding box precision, $1.0$pp on $AP_r$, $0.7$pp on $AP_c$ and $0.9$pp on $AP_f$ compared to SE-R50-GOL. Compared to GOL with SE-R101,  APA* with AGLU also improves the performance by $1.0$pp on $AP$, $0.6$pp on $AP^r$, $1.4$pp on $AP^c$, $0.7$pp on $AP^f$ and $1.1$pp on $AP^b$. This highlights that our APA* and AGLU modules are robust for the rare classes and they outperform the previous SOTA in long-tailed instance segmentation.
\begin{figure}[t]
    \centering
    \includegraphics[width=1\linewidth]{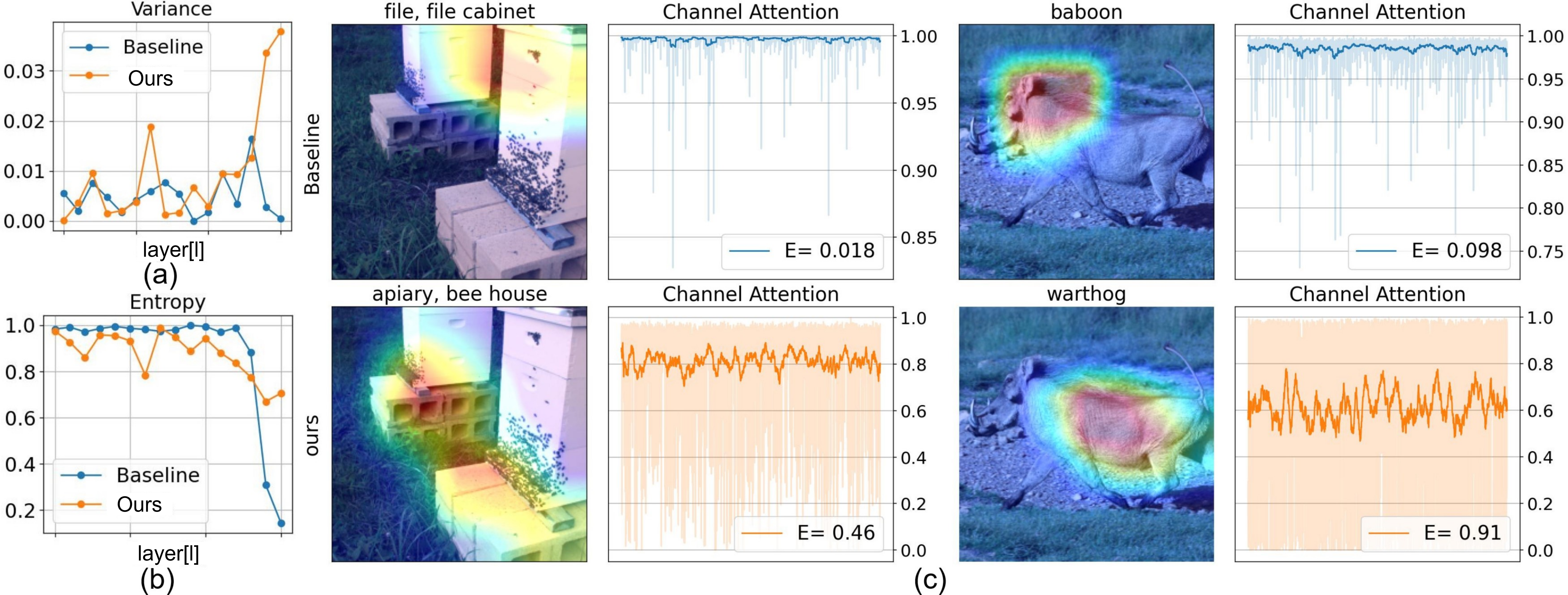}
    \caption{a) APA* (orange curve) increases the attention variance by $0.04$ in the most semantic layer compared to the baseline (blue curve) and removes frequent category attention bias. b) APA* increases the attention entropy in the most semantic layer and retrieves rare class descriptors more efficiently than the baseline. c) Compared to the baseline (top), APA* produces larger entropy attention signals and makes correct rare class predictions. More visualisations are shown in Appendix.}
    \label{fig:qualitative_results}
\end{figure}

\subsection{Qualitative Results.}
We use APA* and Imagenet-LT for our qualitative analysis.
In Fig. \ref{fig:qualitative_results}-(a) we show that APA* increases the variance of channel attention for the most semantic layer by $0.04$ compared to the baseline, making it diverse and informative for all channels. In Fig. \ref{fig:qualitative_results}-(b), we show that APA* increases the entropy of the attention signal for the deeper layers by $0.4\sim0.6$ compared to the baseline, showing that our module produces informative signals that effectively attend to the rare classes.
In Fig.-c \ref{fig:qualitative_results}-top, we show that the baseline channel attention of the rare classes, produces all-pass filters in other words, attention signals that have small entropy, i.e., $E=0.018$ for the \textit{bee house} and  $E=0.098$ for the \textit{warthog} respectively. This hinders rare class learning  and it results in misclassification.  In contrast, our method in Fig.~\ref{fig:qualitative_results}-bottom produces informative channel attention signals that have larger entropy i.e. $E=0.46$ and $E=0.91$ for the \textit{bee house} and \textit{warthog} classes respectively, allowing the model to retrieve rare class features and to make correct predictions. Visualisations  of the learned $\kappa$ and $\lambda$ parameters are provided in the Appendix.

\section{Results on other tasks}
In this section, we comprehensively evaluate the generalisation ability of AGLU activation beyond long-tail learning. We test AGLU in balanced classification, object detection, large-scale supervised pretraining and finetuning, Multi-modal LLMs, image generation with diffusion transformer and pure language models. In every subsection, we introduce the model architecture, training details, dataset and discuss the results.

\subsection{Image Classification}
We perform experiments on ImageNet1K \cite{deng2009imagenet}, using ResNets \cite{he2016deep} and ViT \cite{dosovitskiy2021an} models. We train Resnets with channel attention \cite{hu2018squeeze}, and spatial attention \cite{woo2018cbam} for 100 epochs using learning rate 0.256, batch size 256, weight decay 6e-6, mixup coefficient 0.2, AutoAugment \cite{cubuk2019autoaugment}, training crop size 224 and random-resized-crops. We adopt ViT-S and B variants using a convolutional patch embedding \cite{xiao2021early}, layer-scale \cite{touvron2021going}, cosine positional embeddings and average pooling instead of class token \cite{beyer2022better}. We train these models for 300 epochs using learning rate 0.002, weight decay 0.05,  mixup coefficient 0.2, cutmix coefficient 1.0, batch size 2048, RandAugment \cite{cubuk2020randaugment}, repeated augmentations, color jittering, training crop size 192, random-resized-crops. For ViT-B, we add stochastic depth \cite{huang2016deep} 0.2 to prevent overfitting.

As Table \ref{tab:imagenet1k}-a shows,  our method increases the accuracy for R50, SE-R50 and CBAM-R50, by $0.6$pp, $1.2$pp and $0.6$pp respectively. Also, it increases accuracy by $0.9$pp on SE-R101 and $0.5$pp on SE-R152 showing that AGLU is robust for any model size. In Table  \ref{tab:imagenet1k}-b we show that AGLU has competitive performance for ViT models, increasing by $0.2$pp and $0.3$pp the ViT-S and ViT-B models respectively.

\input{tables/imagenet1k}

\begin{table}[b]
    \centering
    \caption{Dataset details of downstream tasks.}
    \begin{tabular}{c|c|c|c}
    \hline
         Dataset&Train size&Test size& Number of Classes  \\
         \hline
         ImageNet\cite{deng2009imagenet}&1,281,167&50,000&1000\\
         iNaturalist2017 \cite{van2018inaturalist}&579,184&95,986&5,089 \\
         iNaturalist2018 \cite{van2018inaturalist}&437,513&24,426&8,142 \\
         iNaturalist2021 \cite{van2018inaturalist}&2,686,843&500,000	&10,000 \\
         CIFAR10 \cite{krizhevsky2009learning}&50,000&10,000&10 \\
         CIFAR100 \cite{krizhevsky2009learning}&50,000&10,000&100 \\
         Places365 \cite{zhou2014learning}&1,803,459&36,499&365 \\
         
        \hline
    \end{tabular}
    
    \label{tab:downstream_datasets}
\end{table}

\noindent \textbf{Training on ImageNet21K.}
We test AGLU under large scale training by performing ImageNet21k pretraining and finetuning. For this experiment we use a SE-ResNet50 and a simple ViT-B architecture that strictly follows \cite{beyer2022better}. We train these models using batch size 2048, learning rate 0.004, Adam with momentum \cite{kingma2014adam}, weight decay 0.02, mixup coefficient 0.2, training crop size 224, RandAugment and random-resized-crop for 90 epochs. After pretraining, we perform finetuning on 7 downstream tasks, including finegrained classification, scene classification and object classification, using the datasets in Table \ref{tab:downstream_datasets}.

In more detail, we finetune the models for 50 epochs with learning rate 0.04, SGD, no weight decay, batch size 512, mixup coefficient 0.2, cutmix coefficient 1.0, RandAugment, color-jittering, label-smoothing 0.1, train-crop 224 and random-resized-crops.

\input{tables/imagenet21k}

The results are shown in Table \ref{tab:inet21k_down}. Our activation significantly enhances the performance of SE-ResNet50 by $1.9$pp macro-average. AGLU, also increases the performance of VIT-B by $0.2$pp macro average, showing that AGLU is a good activation function for both Transformer and CNN architectures.

\subsection{Experiments with Object detection.}
We perform experiments on COCO \cite{lin2014microsoft}, Objects365v2\cite{shao2019objects365} and the recently proposed V3Det\cite{wang2023v3det}, which is a challenging large scale detection dataset with 13K classes and 243K images.
We report $AP^b$ and $AP^m$ for COCO and $AP^b$ for V3Det and Objects365v2, using primarily SE-ResNets backbones pretrained on ImageNet1K. We use the mmdetection framework and train all models using 2x schedule. 

In Table \ref{tab:object_detection}-a, we show that APA*+AGLU increases the COCO performance of SE-MaskRCNN-R50 by $0.7$pp on $AP^b$ and $AP^m$. Also, it increases the performance of SE-MaskRCNN-R101 by $0.9$pp on $AP^b$ and $0.5$pp on $AP^m$. As Table \ref{tab:object_detection}-b shows, APA*+AGLU improves $AP^b$ by $1.1$pp and $1.6$pp on the vast Objects365v2 dataset using the SE-FasterRCNN-R50 and R101 models respectively. In Table \ref{tab:object_detection}-c, we show that APA*+AGLU increase the performance of SE-FasterRCNN-R50 by $2.9$pp and the performance of SE-Cascade-RCNN-R50 by $2.1$pp on the challenging V3Det, which are significant increases considering that this dataset has 13K classes.
\input{tables/detection_tasks}

\subsection{Experiments with Vision Language Model.} We further test the performance of our proposed activation function in the visual instruction finetuning task using vision-language models (VLMs), following the LLAVA series paradigm \cite{li2024llava,liu2023llava,liu2024llavanext,li2024llavanext-ablations,li2024llavanext-strong}. In more detail, we use an MLP with AGLU activation as a connector module and pretrain two VLMs, including the Qwen2~\cite{yang2024qwen2technicalreport} and Deepstack  \cite{meng2024deepstack} with the SigLIP-so400M \cite{zhai2023sigmoid} vision encoder, using the LLaVA-1.5-PT dataset that has 558k image-text pairs \cite{liu2023improvedllava}, keeping all parameters frozen except for the connector MLP. Next, we perform instruction tuning and we fully finetune the models using the OpenLLaVA-Next dataset \cite{chen2024open}. To evaluate the performance, we use the LMM-evaluation kit  \cite{zhang2024lmmsevalrealitycheckevaluation} that includes many vision-language benchmarks such as MME \cite{fu2023mme}, AI2D \cite{kembhavi2016diagram}, DocVQA \cite{mathew2021docvqa}, GQA \cite{hudson2019gqa}, MMB \cite{liu2024mmbench}, OCRB \cite{Liu_2024_ocrbench}, MMMU \cite{yue2024mmmu}, OKVQA \cite{marino2019ok}, POPE \cite{li2023evaluating}, SQA \cite{lu2022learn}, TextVQA \cite{singh2019towards} and VisWiz \cite{gurari2018vizwiz}. As shown in Table \ref{tab:qwen_bench}, our AGLU connector increases the performance, in both Qwen2-1.5B, Deepstack-7B and Qwen2-7B models by $2.4$pp, $0.6$pp and $0.9$pp, respectively, on the macro average metric and achieves superior performance compared to open source models like LLaVA-Next-7B \cite{liu2024llavanext} and proprietary models like DeepSeek-Janus-Pro 7B. This indicates that AGLU is particularly useful in vision language training because its adaptive behaviour can bridge the domain gap between language and image distributions.
\input{tables/qwen}

\subsection{Experiments in Image Generation.} We try our AGLU activation in the task of image generation. Specifically, we select LightningDIT-XL \cite{yao2025reconstruction}, which is a strong transformer-based, latent diffusion model as our baseline, and we replace all its activation functions with AGLU, keeping any other training details and hyperparameters fixed. We train the AGLU-LightiningDIT-XL model with patch size one, on the Imagenet1k-256 generation benchmark for 64 epochs and evaluate the generation quality using the ADM evaluation suite \cite{dhariwal2021diffusion}. As Table \ref{tab:dit} shows our activation function achieves good performance, decreasing the FID score by $0.1$ and increasing the Recall by $0.8$pp compared to LightningDiT baseline. 
\input{tables/dit}

\subsection{Experiments with Language Model.} 
We perform a language next-token prediction experiment using GPT2~\cite{radford2019language} and the FineWeb-Edu \cite{lozhkov2024fineweb-edu} subset that contains 10 billion GPT2 tokens. The model implementation follows nanogpt by \cite{Karpathy2024}. We train four GPT2 models, GPT2-small that has 124 million parameters, GPT2-medium that has 350 million, GPT-large that has 774 million parameters and GPT-xl that has 1.5 billion parameters for 10 billion tokens, with a batch size of 0.5 million tokens, learning rate $6e-4$ and Adam optimizer with momentum \cite{kingma2014adam}.
We test the model on the HellaSwag benchmark \cite{zellers2019hellaswag} using zero-shot evaluation. To apply AGLU with GPT2, we simply replace all GELU activations with AGLU inside all MLP layers of the transformer. As the results suggest in Table XII, AGLU increases the performance of GPT2 by 0.2pp, 0.4pp, 0.3pp and 0.8pp on the HellaSwag benchmark using GPT2-small, medium, large and xl respectively, and reduces the validation loss on the FineWeb-Edu by 0.01 and 0.02 on GPT2-medium, and GPT2-large and xl respectively showing that AGLU is a good alternative for the next token prediction task, especially for deeper models.
\input{tables/gpt}

\section{Analysis}
\subsection{Computation cost.}
AGLU costs more computing resources compared to non-parametric activation functions because it requires training. The additional computation depends on the architecture, the optimizer and the size of the layer where AGLU is applied.  In Table \ref{tab:aglu_computing_cost}, we compare AGLU against the vanilla architecture for different models, in terms of the number of parameters, peak memory measured in MB, throughput measured in samples per second and FLOPS using ImageNet1K. AGLU requires $17\%$ more memory, it has $3\%$ smaller throughput and it has the same number of parameters and FLOPS in the SE-ResNet50 architecture. In the ViT-B architecture, it lowers the throughput by $3\%$ but keeps all other measures the same.
The extra computing occurs mostly during training and it depends on the network's architecture, i.e., if AGLU is applied after a small linear layer, then the training cost is low, however, if it is applied after a large layer, i.e. a large convolution layer, then the training cost is larger. 

\subsection{Learned activations.} We visualise the converged AGLU activation functions, i.e. the output $AGLU(z,\lambda,\kappa)=y$ for the ImageNet21K-ViT-B and GPT models. To better understand their behaviour, we measure AGLU's expected output $E[y]$, for random inputs $z$ drawn from the Normal distribution. It is striking to notice that both the ViT and GPT2, have the same activation patterns, i.e. the activation of the early layers is harder, i.e., more non-linear and it allows mostly positive inputs to pass whereas in the deeper layers it becomes smoother, i.e. more linear, allowing both positive and negative outputs to propagate.  We hypothesize that, given the models' hierarchical feature extraction ability, the network distinguishes the early features easily and draw harder decision boundaries with the activation function, but as the layers grow deeper, the features become more abstract and the network distinguishes the features less easily and uses smoother decision boundaries. 

\begin{figure}
    \centering
    \includegraphics[width=0.75\linewidth]{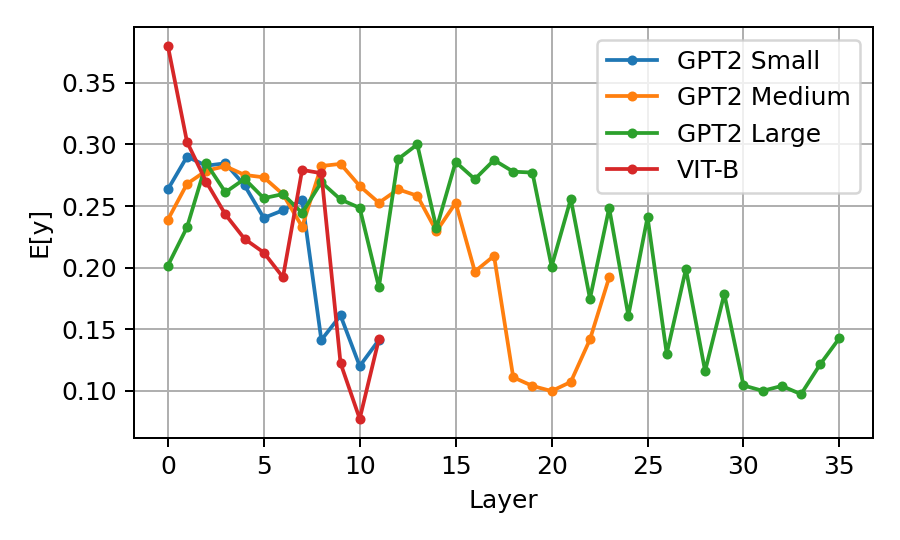}
    \caption{Comparison of the expected AGLU-outputs $E[y]$ for ViT and GPT models, given input drawn from the Normal distribution, across layers. The expected value is larger at the first layers, indicating harder activations, and it becomes smaller at the last layers, indicating smoother activations, for both GPT and ViT. }
    \label{fig:activations_gpt_vit}
\end{figure}
\input{tables/computation_cost}

%% file: tables/imagenet_lt_sota.tex
\begin{table}[t]
\centering
\caption{Top-1 accuracy (\%) on ImageNet-LT test set. E denotes ensemble.}
\resizebox{1\linewidth}{!}{%
    \begin{tabular}{l|c|lll|l}
    
    \bottomrule
    \multicolumn{1}{l|}{Method}             &Backbone& Many & Medium   & Few  & Average  \\ \hline
    MiSLAS \cite{zhong2021improving}        &\multirow{6}{*}{R50}& 61.7& 51.3&35.8& 52.7 \\
    KCL~\cite{kang2021exploring}            &        & 61.8 & 49.4   & 30.9 & 51.5  \\
    TSC \cite{li2022targeted}               &        & 63.5 & 49.7   & 30.4 & 52.4 \\
    RIDE (3E)+CMO\cite{park2022majority}     &        & \underline{66.4} & \underline{53.9}   & 35.6 & \underline{56.2} \\
    DOC \cite{wang2022towards}              &        & 65.1 & 52.8   & 34.2 & 55.0 \\
    CC-SAM \cite{zhou2023class}             &        & 61.4 & 49.5   & \underline{37.1} & 52.4 \\
    \hline
    Our Baseline &\multirow{3}{*}{SE-R50}& 66.2 & 53.1   & \underline{37.1} & 56.0 \\
    APA* (ours)                            &        &67.5 & 54.3& 39.3 &57.4 \\
    APA* + AGLU (ours)                    &        &\textbf{68.3}$^\textbf{+1.9}$ & \textbf{54.8}$^\textbf{+0.9}$& \textbf{39.4}$^\textbf{+2.1}$ &\textbf{57.9}$^\textbf{+1.7}$ \\ 
    \hline
    RIDE (4E) \cite{wang2021long}            &\multirow{9}{*}{X50}        & \underline{68.2} & 53.8   & 36.0 & 56.8 \\
    SSD~\cite{li2021self}                   &        & 66.8 & 53.1   & 35.4 & 56.0 \\
    BCL \cite{zhu2022balanced}              &        & 67.9 & 54.2   & 36.6 & 57.1 \\
    CNT ~\cite{parisot2022long}             &        & 63.2 & 52.1   & 36.9 & 54.2\\
    ALA ~\cite{zhao2022adaptive}            &        & 64.1 & 49.9   & 34.7 & 53.3\\
    ResLT ~\cite{cui2022reslt}              &        & 63.6 &\textbf{55.7}  &\underline{38.9}&56.1\\
    ABC-Norm~\cite{hsu2023abc}              &        & 60.7 & 49.7  & 33.1    &51.7 \\
    RIDE (3E)+CMO+CR~\cite{ma2023curvature} &         & 67.3 & 54.6   & 38.4 & \underline{57.4}\\
    LWS+ImbSAM \cite{zhou2023imbsam}        &        & 63.2 & 53.7   & 38.3 & 55.3 \\ 
    \hline
    Our Baseline      &\multirow{3}{*}{SE-X50}& 67.9 & 53.0   & 37.7 & 56.7 \\
    APA* (ours)                             &        &68.9& 55.4& 39.4&58.4 \\
    APA* + AGLU (ours)                             &        &\textbf{69.8}$^\textbf{+1.6}$& \underline{55.7}$^\text{0.0}$& \textbf{41.1}$^\textbf{+2.2}$&\textbf{59.1}$^\textbf{+1.7}$ \\
    \bottomrule
    \end{tabular}
}
\label{tab:imagenet_sota}
\end{table}

%% file: tables/inat_places_cifar_sota.tex
\begin{table}[t]
\centering
\caption{(a) Results of APA* and AGLU for iNaturalist and Places-LT. (b) Results of APA and AGLU for CIFAR100-LT with imbalance 10 and 100. }
\vspace*{-1em}
\begin{tabular}{cc}
\resizebox{0.465\linewidth}{!}{%
\begin{tabular}{l|c|l}
    \toprule
         Method&iNat18&PlacesLT \\
         \hline
         DisAlign~\cite{zhang2021distribution}&70.6&39.3\\
         MisLAS~\cite{zhong2021improving}& 71.6&40.4\\
         LADE~\cite{hong2021disentangling}&70.0&38.8\\
         ALA~\cite{zhao2022adaptive}&70.7&40.1\\
         TSC~\cite{li2022targeted}&69.7&-\\
         CNT~\cite{parisot2022long}&-&39.2\\
         WD+MaxNorm~\cite{alshammari2022long}&70.2&-\\
         DOC~\cite{wang2022towards}&71.0&-\\
         BCL\cite{zhu2022balanced} & \underline{71.8} &-\\
         ResLT~\cite{cui2022reslt}&70.5&39.8\\
         IIF~\cite{alexandridis2023inverse}&-&40.2\\
         ABC Norm~\cite{hsu2023abc}&71.4&-\\
         
         LWS+ImbSAM\cite{zhou2023imbsam}&71.1&-\\
         CC-SAM\cite{zhou2023class}& 70.9&\underline{40.6}\\
         AREA \cite{chen2023area}&68.4 &-\\
         \hline
         Baseline with SE&71.3&40.5\\
         APA* (ours)&72.3&41.3\\
         APA* +AGLU (ours)&\textbf{74.8}$^\textbf{+3.0}$&\textbf{42.0}$^\textbf{+1.4}$\\
    \bottomrule
    \end{tabular}
}
&
\resizebox{0.465\linewidth}{!}{%
\begin{tabular}{l|ll}
    \toprule
         \multirow{ 1}{*}{Method}&10&100\\
         \hline
         BALMS \cite{Ren2020balms}&63.0&50.8\\
         RIDE (4E) \cite{wang2021long}&-&49.4\\
         ACE (4E)~\cite{cai2021ace}&-&49.6\\
         DiVE~\cite{he2021distilling}&62.0&45.4\\
         SSD~\cite{li2021self}&62.3&46.0\\
         MisLas \cite{zhong2021improving}&63.2&47.0\\
         HSC~\cite{wang2021contrastive}&63.1&46.7\\
         LADE \cite{hong2021disentangling}&61.7&45.4\\
         ResLT (3E)~\cite{cui2022reslt}&63.7&49.7\\
         TLC (4E) \cite{li2022trustworthy}&-&49.8\\
         TSC \cite{li2022targeted}&59.0&43.8\\
         RIDE+CMO~\cite{park2022majority}&60.2& 50.0\\
         CC-SAM~\cite{zhou2023class}&- &50.8\\
         RIDE+CMO+CR\cite{ma2023curvature}&61.4&50.7\\
         AREA \cite{chen2023area}& 60.8 & 48.9\\
         \hline
         Baseline with SE&\underline{65.2}&\underline{50.9}\\
         APA* (ours)&65.7&51.9\\
         APA*+AGLU(ours)&\textbf{66.8}$^\textbf{+1.6}$&\textbf{52.3}$^\textbf{+1.4}$\\
         
    \bottomrule    
    \end{tabular}
}
\\\specialcell[t]{\small(a)}&\specialcell[t]{\small(b)}
\end{tabular}

\label{tab:inat_places}

\end{table}

%% file: tables/ablations.tex
\begin{table*}[t]
\centering
\caption{For all ablations we use ImageNet-LT. In (a) we compare adaptive activation to other learnable activation functions using SE-ResNet50. In (b) we show that adaptive activation and AGLU generalise to other attention mechanisms. In (c) we show that APA* can be combined with classifier learning techniques. In (d) we show that APA* works effectively with deeper ResNets. In (e) we show the components of APA*. In (f) we compare AGLU with other activation functions. }
\vspace*{-1em}

\begin{tabular}{ccc}
\resizebox{0.23\linewidth}{!}{%
\begin{tabular}{c|ccc|c}
    \toprule
         Activation&Many&Med.&Few&Avg\\
         \hline
         Sigmoid&66.2 & 53.1   & 37.1 & 56.0\\
         with Temp& 65.9 &\textbf{53.8}&\textbf{40.3}&56.6\\
         \hline
         Gumbel&66.2& 53.2 &39.3&56.3\\
         with Temp& \underline{66.9} &\underline{53.4}&\underline{39.7}&56.7\\
         \hdashline 
         APA&\textbf{67.1} & \textbf{53.8}& 39.6 &\textbf{57.0}\\
    \bottomrule
    \end{tabular}%
}
&
\resizebox{0.2\linewidth}{!}{%
\begin{tabular}{c|c}
    \toprule
        Attention type&Avg\\
         \hline
         Spatial~\cite{woo2018cbam}&54.8\\
         +APA&55.2\\
         +APA + AGLU&\textbf{56.4}\\ 
         \hline
         Spatial + Channel ~\cite{woo2018cbam}&55.6\\
         +APA &56.9\\
        +APA + AGLU &\textbf{57.1}\\
         
        \toprule
    \end{tabular}
}&
\resizebox{0.24\linewidth}{!}{
\begin{tabular}{c|ccc}
    \hline
         Method&SE&APA* \\
         \hline
         CE &51.7&\textbf{52.9}\\
         PC-Softmax \cite{hong2021disentangling}&56.0&\textbf{57.4}\\
         cRT \cite{kang2019decoupling} &55.6&\textbf{56.4}\\
         Decoupled-DRW ~\cite{cao2019learning}&55.1&\textbf{56.6}\\
         BSCE~\cite{Ren2020balms}&56.0&\textbf{57.0}\\
         \hline
    \end{tabular}
}
\\\specialcell[t]{\small(a) APA comparison.}&\specialcell[t]{\small(b) Attention types w/ R50.}&\specialcell[t]{\small(c) Classifier learning.}\\

\resizebox{0.25\linewidth}{!}{%
\begin{tabular}{c|ccc|c}
    \toprule
         Method&Many&Med.&Few&Avg\\
         \hline
         SE-R101 &67.5&53.3&37.6&56.7\\
         \hdashline 
         APA*-R101&\textbf{68.1} & \textbf{56.0}& \textbf{42.1} &\textbf{58.8}\\
        \hline
        SE-R152&68.0&54.4&39.8&57.6\\
        \hdashline 
         APA*-R152&\textbf{69.0} & \textbf{56.8}& \textbf{41.2} &\textbf{59.4}\\
    \bottomrule
    \end{tabular}%
}
&
\resizebox{0.28\linewidth}{!}{%
\begin{tabular}{c|c|c|c|c}
    \hline
    SE&APA&Dropout&LayerNorm&Avg  \\
    \hline
    &&&&55.0\\
    \checkmark&&&&56.0\\
    \checkmark&\checkmark&&&57.0\\
    \checkmark&\checkmark&\checkmark&&57.3\\
    \checkmark&\checkmark&\checkmark&\checkmark&\textbf{57.4}\\
    \hline
    \end{tabular}
}&
\resizebox{0.15\linewidth}{!}{
\begin{tabular}{c|c}
    \toprule
         Activations&Avg \\
         \hline
         ReLU&57.4\\
         PReLU \cite{he2015delving} &54.8\\
         ELU \cite{clevert2015fast}&52.6\\
         Mish \cite{misra2019mish} &57.4\\
         GELU \cite{hendrycks2016gaussian}&57.5\\
         SiLU \cite{hendrycks2016gaussian}&57.1\\
         AGLU&\textbf{57.9}\\
    \toprule
    \end{tabular}
}
\\\specialcell[t]{\small(d)Deeper Networks.}&\specialcell[t]{\small(e) APA* ablation.}&\specialcell[t]{\small(f) AGLU comparison.}\\
\end{tabular}

\label{tab:agca_ablations}

\end{table*}

%% file: tables/lvis.tex
\begin{table}[t]
    \centering
    \caption{Comparisons on LVISv1.0 using MaskRCNN-FPN and 2x schedule.}  
    \vspace*{-1em}
    \resizebox{1\linewidth}{!}{%
        \begin{tabular}{c|c|lllll}
        \toprule
         Method&Backbone&$AP^m$&$AP^r$&$AP^c$&$AP^f$&$AP^b$ \\
         \hline
          RFS ~\cite{gupta2019lvis}&\multirow{8}{*}{R50} &23.7&13.3&23.0&29.0&24.7\\
          IIF~\cite{alexandridis2023inverse}& & 26.3& 18.6& 25.2& 30.8& 25.8\\
          Seesaw~\cite{wang2021seesaw}& &26.4&19.6&26.1&29.8&27.4\\
          LOCE \cite{feng2021exploring}&  &26.6 &18.5 &26.2&30.7&27.4\\
          PCB+Seesaw~\cite{he2022relieving}&&27.2&19.0&27.1&30.9&\underline{28.1}\\
          ECM ~\cite{hyun2022long}&&27.4&19.7&27.0&\underline{31.1}&27.9\\
          GOL ~\cite{alexandridis2022long}& &27.7&\underline{21.4}&27.7&30.4&27.5\\
          ECM+GAP~\cite{zhang2023reconciling}& &26.9&20.1&26.8&30.0&27.2\\
        \hline
         GOL (baseline)& SE-R50&\underline{28.2}&20.6&\underline{28.9}&30.8&\underline{28.1}\\
         GOL+AGLU(ours)& APA*-R50 &\textbf{29.1}$^\textbf{+0.9}$&\textbf{21.6}$^\textbf{+0.2}$&\textbf{29.6}$^\textbf{+0.7}$&\textbf{31.7}$^\textbf{+0.6}$&\textbf{29.0}$^\textbf{+0.9}$\\
         \hline
         RFS \cite{gupta2019lvis}&\multirow{8}{*}{R101~\cite{he2016deep}}&27.0&16.8&26.5&32.0&27.3\\
         NorCal \cite{pan2021model}& &27.3&20.8&26.5&31.0&28.1\\
         Seesaw \cite{wang2021seesaw}& &28.1&20.0&28.0&31.8&28.9\\
         GOL \cite{alexandridis2022long}& &29.0&22.8&29.0&31.7&29.2\\
         ECM \cite{hyun2022long}& &28.7&21.9&27.9&32.3&29.4\\
         PCB + Seesaw\cite{he2022relieving}& &28.8&22.6&28.3&32.0&29.9\\ 
         ROG \cite{zhang2023reconciling}& &28.8&21.1&29.1&31.8&28.8\\
         \hline
         GOL (baseline)& SE-R101&\underline{29.7}&\underline{23.0}&\underline{29.9}&\underline{32.5}&\underline{30.0}\\
         GOL+AGLU(ours)& APA*-R101 &\textbf{30.7}$^\textbf{+1.0}$&\textbf{23.6}$^\textbf{+0.6}$&\textbf{31.3}$^\textbf{+1.4}$&\textbf{33.1}$^\textbf{+0.7}$&\textbf{31.1}$^\textbf{1.1}$\\
         \bottomrule
    \end{tabular}
    }
    
    \label{tab:lvis1_sota}
\end{table}

%% file: tables/imagenet1k.tex
\begin{table}[t]
\centering
\caption{Results on ImageNet1K using ResNets in (a) and vision Transformers in (b).}
\begin{tabular}{cc}
\resizebox{0.5\linewidth}{!}{%

 \begin{tabular}{c|c}
    \hline
         Method&top-1  \\
         \hline
         ResNet50 \cite{hu2018squeeze}&76.9\\
         w/ AGLU &\textbf{77.5}\\
         \hline
         SE-ResNet50 \cite{hu2018squeeze}&77.5\\
         w/ APA*+AGLU&\textbf{78.7}\\
         \hline
         CBAM-ResNet50 \cite{hu2018squeeze}&78.3\\
         w/ APA*+AGLU &\textbf{78.9}\\
          \hline
         SE-ResNet101&79.4\\
         w/ APA* + AGLU&\textbf{80.3}\\
         \hline
         SE-ResNet152&80.3\\
         w/ APA* + AGLU&\textbf{80.8}\\
         \bottomrule
    \end{tabular}
}
&
\resizebox{0.35\linewidth}{!}{
 \begin{tabular}{c|c}
    \hline
         Method&top-1  \\
         \hline
         ViT-S &80.2\\
         w/ AGLU &\textbf{80.4}\\
         \hline
         ViT-B&82.5\\
         w/ AGLU&\textbf{82.8}\\
         \bottomrule
    \end{tabular}

}
\\\specialcell[t]{\small(a) Using ResNets}&\specialcell[t]{\small(b) Using ViT}\\

\end{tabular}

\label{tab:imagenet1k}
\end{table}

%% file: tables/imagenet21k.tex
\begin{table*}[t]
    \centering
    \caption{Comparative downstream results on 7 popular image classification datasets using 224 training and validation crops.}
        \begin{tabular}{c|c|c|c|c|c|c|c|c}
        \hline
        Method&Imagenet1K&iNaturalist 2017&iNaturalist 2018&iNaturalist 2021&CIFAR10&CIFAR100&Places365&Average \\ 
        \hline
        SE-Resnet50&80.5&67.3&73.4&82.1&96.0&86.4&54.3&77.1\\
        w/ APA* + AGLU&\textbf{81.5}&\textbf{69.7}&\textbf{76.4}&\textbf{83.9}&\textbf{97.6}&\textbf{87.6}&\textbf{54.7}&\textbf{78.8}\\
        \hline
        ViT-B &83.9&72.1&78.9&84.7&99.1&93.5&56.9&81.3\\
        w/ AGLU &\textbf{84.0}&\textbf{72.2}&\textbf{79.3}&\textbf{85.1}&\textbf{99.2}&\textbf{93.6}&\textbf{57.1}&\textbf{81.5}\\
        \hline
        \end{tabular}
    \label{tab:inet21k_down}
\end{table*}

%% file: tables/detection_tasks.tex
\begin{table*}[t]
\centering
\caption{Resuls on balanced tasks. In (a) and (b) and (c), we combine APA*+AGLU on COCO, Objects365v2 and V3Det dataset respectively. In (d) we combine AGLU and APA* to ResNet, SE-ResNet and CBAM-ResNet in ImageNet1K. The $^\dagger$ shows that the baseline is taken from mmdetection \cite{mmdetection}.}
\vspace*{-1em}
\begin{tabular}{ccc}

\resizebox{0.35\linewidth}{!}{%

\begin{tabular}{c|c|c}
    \hline
         Method&$AP^{b}$&$AP^{m}$\\
         \hline
         MaskRCNN-ResNet50&39.2&35.4\\
         SE-MaskRCNN-ResNet50&40.5&36.9\\
         w/ APA*+AGLU&\textbf{41.2}&\textbf{37.6}\\
         \hline
         SE-MaskRCNN-ResNet101&42.7&38.4\\
         w/ APA*+AGLU&\textbf{43.6}&\textbf{38.9}\\
         \hline
    \end{tabular}
}
&
\resizebox{0.3\linewidth}{!}{%

\begin{tabular}{c|c}
    \hline
         Method&$AP^{b}$\\
         \hline
         FasterRCNN-ResNet50$^\dagger$&19.8\\
         SE-FasterRCNN-ResNet50&26.4\\
         w/ APA*+AGLU&\textbf{27.5}\\
         \hline
         SE-FasterRCNN-ResNet101&29.3\\
         w/ APA*+AGLU&\textbf{30.9}\\
         \hline
    \end{tabular}
}
&
\resizebox{0.25\linewidth}{!}{%

\begin{tabular}{c|c}
    \hline
         Method&$AP^{b}$\\
         \hline
         FasterRCNN-Resnet50&25.4\\
         w/ SE&27.0\\
         w/ APA*+AGLU&\textbf{29.9}\\
        \hline
         CascadeRCNN-ResNet50&31.6\\
         w/ SE&33.3\\
         w/ APA*+AGLU&\textbf{35.4}\\
        \bottomrule
    \end{tabular}
}
\\\specialcell[t]{\small(a) COCO}&\specialcell[t]{\small(b) Objects365v2}&\specialcell[t]{\small(c) V3Det}\\

\end{tabular}

\label{tab:object_detection}
\end{table*}

%% file: tables/qwen.tex
\begin{table*}[t]
    \centering
    \caption{Comparative results on the visual instruction following task, using a Qwen2-SigLIP model.}
    \resizebox{1\linewidth}{!}{%
        \begin{tabular}{c|c|c|c|c|c|c|c|c|c|c|c|c|c|c|c|c}
        \toprule
             Method&MME&AI2D&DOCVQA&GQA&MMB$^{\text{CC}}$&MMB$^{\text{CN}}$&MMB$^{\text{EN}}$&MMB$^{\text{RU}}$&MMMU&OCRB&OKVQA&POPE&SQA$^{\text{I}}$&VQA$^{\text{T}}$&VizWiz&AVG  \\
             \hline
             DeepSeek-Janus-Pro-7B \cite{chen2025janus}&1567&-&-&59.1&-&-&79.2&-&41.0&-&-&87.4&-&-&-&-\\
             LLaVA-1.5-7B \cite{liu2023improvedllava}&1511&-&-&62.0&-&58.3&64.3&-&-&-&-&86.0&71.6&61.3&50.0&-\\
             LLaVA-Next-7B \cite{liu2024llavanext}&1519&66.6&74.4&64.2&-&60.6&67.4&-&35.8&-&-&86.5&70.1&64.9&57.6&-\\
             VILA-7B \cite{lin2024vila}&1533&-&-&62.3&-&61.7&70.3&-&-&-&-&85.5&68.3&64.4&57.8&-\\
             InternVL-Chat-7B \cite{chen2024internvl}&1525&-&-&62.9&-&-&-&-&-&-&-&87.1&-&57.0&52.5&-\\
             \hline
             Qwen2-1.5B-SigLIP&\textbf{1393}&59.3&46.6&60.3&33.7&57.3&57.6&33.8&\textbf{27.9}&\textbf{42.5}&51.4&86.5&\textbf{70.3}&53.4&37.5&52.5\\
             w/ AGLU&1390&\textbf{60.3}&\textbf{47.7}&\textbf{60.9}&\textbf{37.7}&\textbf{60.1}&\textbf{64.3}&\textbf{47.6}&26&41.3&\textbf{52.3}&\textbf{87.5}&69.8&\textbf{54.9}&\textbf{43.4}&\textbf{54.9}\\
             \hline
             Deepstack-7B \cite{meng2024deepstack}&\textbf{1592}&74.6&73.2&\textbf{64.4}&46.7&72.8&75.3&51.0&47.1&\textbf{53.7}&56.2&\textbf{88.0}&71.4&66.1&\textbf{61.7}&65.5\\
             w/ AGLU&1573&\textbf{75.3}&\textbf{73.6}&64.3&\textbf{47.3}&\textbf{73.6}&\textbf{76.3}&\textbf{51.3}&\textbf{49.0}&53.0&\textbf{58.0}&87.9&\textbf{76.1}&\textbf{66.4}&61.0&\textbf{66.1}\\
             \hline
             Qwen2-7B-SigLIP&1596&73.9&74.8&63.9&42.9&\textbf{72.3}&74.2&\textbf{56.3}&\textbf{45.8}&55.9&61.2&87.7&70.6&\textbf{68.0}&\textbf{53.8}&65.4\\
             w/ AGLU&\textbf{1634}&\textbf{74.1}&\textbf{75.6}&\textbf{64.3}&\textbf{46.9}&71.8&\textbf{75.1}&55.1&45.7&\textbf{58.3}&\textbf{61.9}&\textbf{88.1}&\textbf{76.5}&\textbf{68.0}&51.0&\textbf{66.3}\\
             \bottomrule
        \end{tabular}
    }
    
    \label{tab:qwen_bench}
\end{table*}

%% file: tables/dit.tex
\begin{table}[htb]
    \centering
    \caption{AGLU achieves good performance in image generation.}
    \resizebox{1\linewidth}{!}{%
        \begin{tabular}{c|ccccc}
        \toprule
             Model&FID&sFID&Inception Score&Precision&Recall  \\
             \hline
             LightiningDIT-XL\cite{yao2025reconstruction}&2.2&4.3&\textbf{251.9}&\textbf{81.5}&58.3\\
             LightiningDIT-XL w/ AGLU&\textbf{2.1}&\textbf{4.2}&251.1&81.0&\textbf{59.1}\\
             \hline
        \end{tabular}
    }
    
    \label{tab:dit}
\end{table}

%% file: tables/gpt.tex
\begin{table}[htb]
    \centering
    \caption{Comparative results using GPT2 model on the HellaSwag benchmark and Fineweb-Edu validation set.}
    \begin{tabular}{c|c|c}
    \hline
    Method&HellaSwag&FineWeb Val-Loss \\ 
    \hline
    GPT2-small&30.8&3.08\\
    GPT2-small with AGLU&\textbf{31.0}&3.08\\
    \hline
    GPT2-medium&35.2&2.86\\
    GPT2-medium with AGLU&\textbf{35.6}&\textbf{2.85}\\
    \hline
    GPT2-large&38.8&2.72\\
    GPT2-large with AGLU&\textbf{39.1}&\textbf{2.70}\\
    \hline
    GPT2-xl&42.7&2.62\\
    GPT2-xl with AGLU&\textbf{43.5}&\textbf{2.60}\\
    \hline
    \end{tabular}
    \label{tab:hella_swag}
\end{table}

%% file: tables/computation_cost.tex
\begin{table}[t]
    \centering
    \caption{Computation cost of AGLU. FLOPS are measured on a V100 GPU using batch-size of 1, $224^2$ image resolution and mixed precision. Peak memory (MB) and throughput ($\frac{\text{img}}{\text{sec}}$) are measured on the ImageNet1K finetuning.}  
        \begin{tabular}{c|c|c|c|c}
        \toprule
         Model&Parameters&MB&$\frac{\text{img}}{\text{sec}}$&GFLOPS\\
         \bottomrule
         SE-R50&28M&\textbf{6020}&\textbf{393.6}&8.2\\
         w/ APA*+AGLU&28M&7064&382.7&8.2\\
         \hline
         ViT-B&86M&5145&\textbf{263.7}&33.5\\
         w/ AGLU&86M&5145&256.9&33.5\\
         \hline     
    \end{tabular}
    \label{tab:aglu_computing_cost}
\end{table}

%% file: sections/conclusion.tex
\section{Discussion}
We have shown that the adaptive activation function is a solid alternative to standard activation functions such as RELU or GELU, for many vision tasks, especially for long-tailed tasks. Some of the biggest benefits are that, it is a generalisation of many common activation functions, it can be computed analytically and it is a easy to use. We believe that AGLU, can be particularly effective when applied to specialised modules such as in the VLM connector, or CNN detectors and is a good alternative for large scale pretrained Transformer models, showing $0.3$pp increase in ViT and GPT models for both classification and generation tasks.  This is lower than the performance increase observed with CNNs, potentially because the transformer is a highly non-linear structure and already very expressive due to its attention mechanism. Regarding, the computational resources, AGLU is overall more expensive than non-adaptive functions but the computing cost is diminishing when AGLU is applied to small linear layers.

\section{Conclusion}
\label{sec:conclusion}
Our work highlights the impact of the activation function inside the model's activations for balanced and imbalanced data distributions. We have empirically shown that the degree of data imbalance affects the logit distributions and the intermediate signals and we have shown that the commonly used Sigmoid activation function is unable to model the intermediate features. To this end, we have proposed an novel adaptive parametric activation that unifies most common activation functions under the same formula and we have tested it in several long-tail and balanced benchmarks showing great generalisation.